\documentclass{article}

\usepackage[final]{neurips_2023}

\usepackage[utf8]{inputenc} 
\usepackage[T1]{fontenc}    
\usepackage{hyperref}       
\usepackage{url}            
\usepackage{booktabs}       
\usepackage{amsfonts}       
\usepackage{nicefrac}       
\usepackage{microtype}      
\usepackage{xcolor}         
\usepackage{algorithm}
\usepackage{algorithmic}
\usepackage{amsmath}
\usepackage{amssymb}
\usepackage{mathtools}
\usepackage{amsthm}
\usepackage{bm}
\usepackage{wrapfig}
\usepackage[capitalize,noabbrev]{cleveref}
\usepackage{natbib}
\usepackage{multirow}
\usepackage{selectp}

\setcitestyle{round}
\hypersetup{
   colorlinks=true,
   linkcolor=[RGB]{30, 30, 180},
   citecolor=[RGB]{30, 30, 180},
   urlcolor=magenta,
   pdfborder=0 0 0,
   pdftitle={},
   pdfsubject={}, 
   pdfkeywords={},
   pdfauthor={},%
   pdfstartview=FitH
}

\usepackage{array}

\newcommand\Ba{\bm{a}}

\newcommand\Bc{\bm{c}}

\newcommand\Be{\bm{e}}

\newcommand\Bh{\bm{h}}

\newcommand\Bo{\bm{o}}

\newcommand\Bs{\bm{s}}

\newcommand\Bv{\bm{v}}

\newcommand\Bx{\bm{x}}
\newcommand\By{\bm{y}}

\newcommand\BC{\bm{C}}

\newcommand\BE{\bm{E}}

\newcommand\BP{\bm{P}}

\newcommand\Bbe{\bm{\beta}}

\newcommand\Bga{\bm{\gamma}}
\newcommand\Bmu{\bm{\mu}}

\newcommand\Bsi{\bm{\sigma}}

 \newcommand{\dR}{\mathbb{R}}

 \newcommand{\cN}{\mathcal{N}}
 \newcommand{\cP}{\mathcal{P}}
 
\newcommand{\cS}{\mathcal{S}} 
 \newcommand{\cV}{\mathcal{V}}

\makeatletter
\newcommand{\dlmf}[1]{%
\citep[%
  \def\nextitem{\def\nextitem{, }}%
  \@for \el:=#1\do{\nextitem\href{http://dlmf.nist.gov/\el}{(\el)}}%
]{olver_nist_2010}%
}
\makeatother

\title{Semantic HELM: A Human-Readable Memory\\ for Reinforcement Learning}

\author{
  \textbf{Fabian Paischer}$~^{1}$,
  \textbf{Thomas Adler}$~^{1}$,
  \textbf{Markus Hofmarcher}$~^{2}$,
  \textbf{Sepp Hochreiter}$~^{1}$ \\
  $~^{1}$~ELLIS Unit Linz and LIT AI Lab, Institute for Machine Learning,\\
  $~^{2}$~JKU LIT SAL eSPML Lab, Institute for Machine Learning,\\
                  ~~~~Johannes Kepler University, Linz, Austria\\
  \texttt{paischer@ml.jku.at}
}

\begin{document}

\maketitle

\begin{abstract}
Reinforcement learning agents deployed in the real world often have to cope with partially observable environments. 
Therefore, most agents employ memory mechanisms to approximate the state of the environment. 
Recently, there have been impressive success stories in mastering partially observable environments, mostly in the realm of computer games like Dota 2, StarCraft II, or MineCraft. 
However, existing methods lack interpretability in the sense that it is not comprehensible for humans what the agent stores in its memory.
In this regard, we propose a novel memory mechanism that represents past events in human language.
Our method uses CLIP to associate visual inputs with language tokens. 
Then we feed these tokens to a pretrained language model that serves the agent as memory and provides it with a coherent and human-readable representation of the past.
We train our memory mechanism on a set of partially observable environments and find that it excels on tasks that require a memory component, while mostly attaining performance on-par with strong baselines on tasks that do not. 
On a challenging continuous recognition task, where memorizing the past is crucial, our memory mechanism converges two orders of magnitude faster than prior methods.
Since our memory mechanism is human-readable, we can peek at an agent's memory and check whether crucial pieces of information have been stored.
This significantly enhances troubleshooting and paves the way toward more interpretable agents.
\end{abstract}

\section{Introduction}
In reinforcement learning (RL) an agent interacts with an environment and learns from feedback provided in the form of a reward function.
In many applications, especially in real-world scenarios, the true state of the environment is not directly accessible to the agent, but rather approximated via observations that reveal mere parts of it. 
In such environments, the capability to approximate the true state by virtue of an agent's perception is crucial \citep{astrom_optimal_1964,kaelbling_planning_1998}.
To this end, many approaches track events that occurred in the past. 
The brute-force strategy is to simply store all past observations. 
However, this is often infeasible and it is much more efficient to store more abstract representations of the history.
Thus, many RL algorithms use memory mechanisms such as LSTM \citep{hochreiter_long_1997} or Transformer \citep{vaswani_attention_2017} to compress sequences of high-dimensional observations.
This has led to impressive successes mostly in the realm of mastering computer games on a human or even super-human level. 
Some examples are Dota 2 \citep{berner_dota_2019}, StarCraft II \citep{vinyals_grandmaster_2019}, or MineCraft \citep{baker_video_2022, patil_align-rudder_2022}. 

Most state-of-the-art methods dealing with partial observability in RL employ a memory mechanism that is not intelligible for humans.
In this regard, we draw inspiration from the semantic memory present in humans \citep{yee_semantic_2017} and propose to represent past events in human language. 
Humans memorize abstract concepts rather than every detail of information they encountered in the past \citep{richards_persistence_2017,bowman_abstract_2018}.
Their ability to abstract is heavily influenced by the exposure to language in early childhood \citep{waxman_words_1995}.
Further, humans use language on a daily basis to abstract and pass on information. 
Therefore, language is a natural choice as a representation for compounding information and has the key advantage of being human-readable.
This enables analyzing if critical pieces of information have entered the memory or not.
Based on this data, it becomes clear which parts of the system require refinement. 
Moreover, natural language has been shown to be effective as a compressed representation of past events in RL \citep{paischer_history_2022}.

\begin{figure}[t]
\centerline{\includegraphics[width=\textwidth]{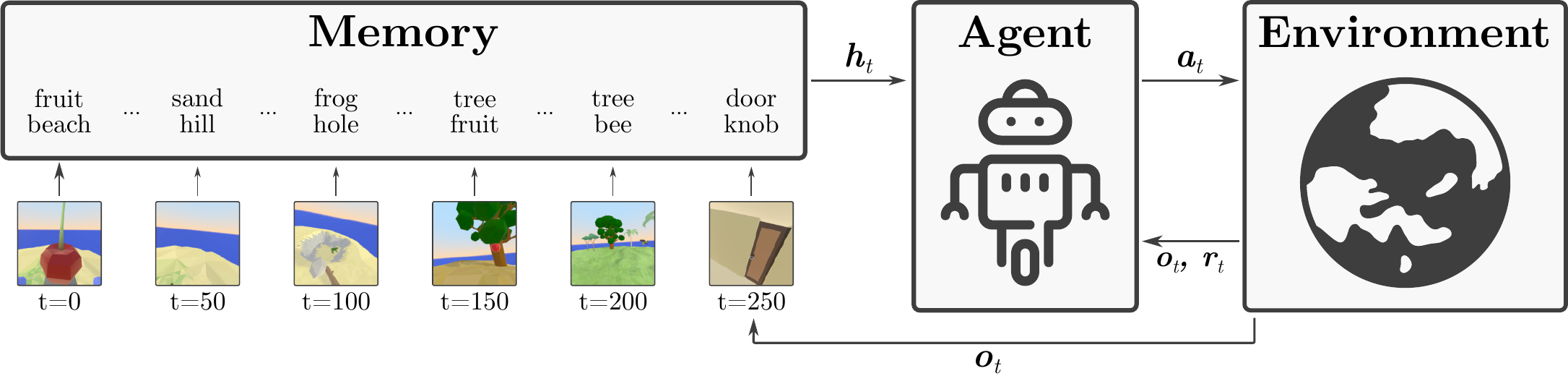}}
\caption{We add a semantic and human-readable memory to an agent to tackle partially observable RL tasks. 
We map visual observations $\Bo_t$ to the language domain via CLIP retrieval. 
The memory component, a pretrained language encoder, operates on text only and compresses a history of tokens into a vector $\Bh_t$. 
The agent takes an action $\Ba_t$ based on the current observation $\Bo_t$ and the compressed history $\Bh_t$.}
\label{fig:overview}
\vskip -0.2in
\end{figure}

Our proposed method, Semantic HELM (SHELM), leverages pre-trained foundation models to construct a memory mechanism that does not require any training.
We use CLIP \citep{radford_learning_2021} to associate visual inputs with language tokens.
Thereby, the vocabulary of the CLIP language encoder serves as a semantic database of concepts from which we retrieve the closest tokens to a given observation.
These tokens are passed to a pretrained language model that serves as memory and provides the agent with a coherent representation of the past. 

We illustrate the benefits of a human-readable and semantic memory in partially observable RL problems.
First, we conduct a qualitative analysis on whether a CLIP vision encoder is capable of extracting semantics out of synthetic environments.
Then, we test SHELM on a set of partially observable 2D MiniGrid \citep{chevalier-boisvert_minimalistic_2018}, and 3D MiniWorld \citep{chevalier-boisvert_miniworld_2018} environments.
We find that even though these environments are only partially observable, they often do not necessitate a memory mechanism as they are solvable by a memory-less policy.
The MiniGrid-Memory task, however, clearly requires memory and SHELM reaches state-of-the-art performance.
On more realistic 3D environments such as Avalon \citep{albrecht_avalon_2022} and Psychlab \citep{leibo_psychlab_2018}, SHELM successfully assesses the semantics of visual observations.
In turn, it requires approximately 100 times fewer interaction steps than prior methods on Psychlab's continuous recognition task that explicitly evaluates for memory capacity.
On Avalon, we find that the addition of our semantic memory performs on-par with the current state-of-the-art for a limited interaction budget, while adding interpretability to the memory.

\section{Methods}
\label{sec:methodology}

Our goal is to express visual observations in language space such that the resulting representations become comprehensible for humans.
To this end, we instantiate a mapping from images to text in the form of pretrained components which are entirely frozen during training.
This way the available computational resources are invested into performance rather than interpretability. 
Before we describe SHELM, we briefly review the HELM framework, which serves as a starting point for our work. 

\begin{figure}[t]
\centerline{\includegraphics[width=\textwidth]{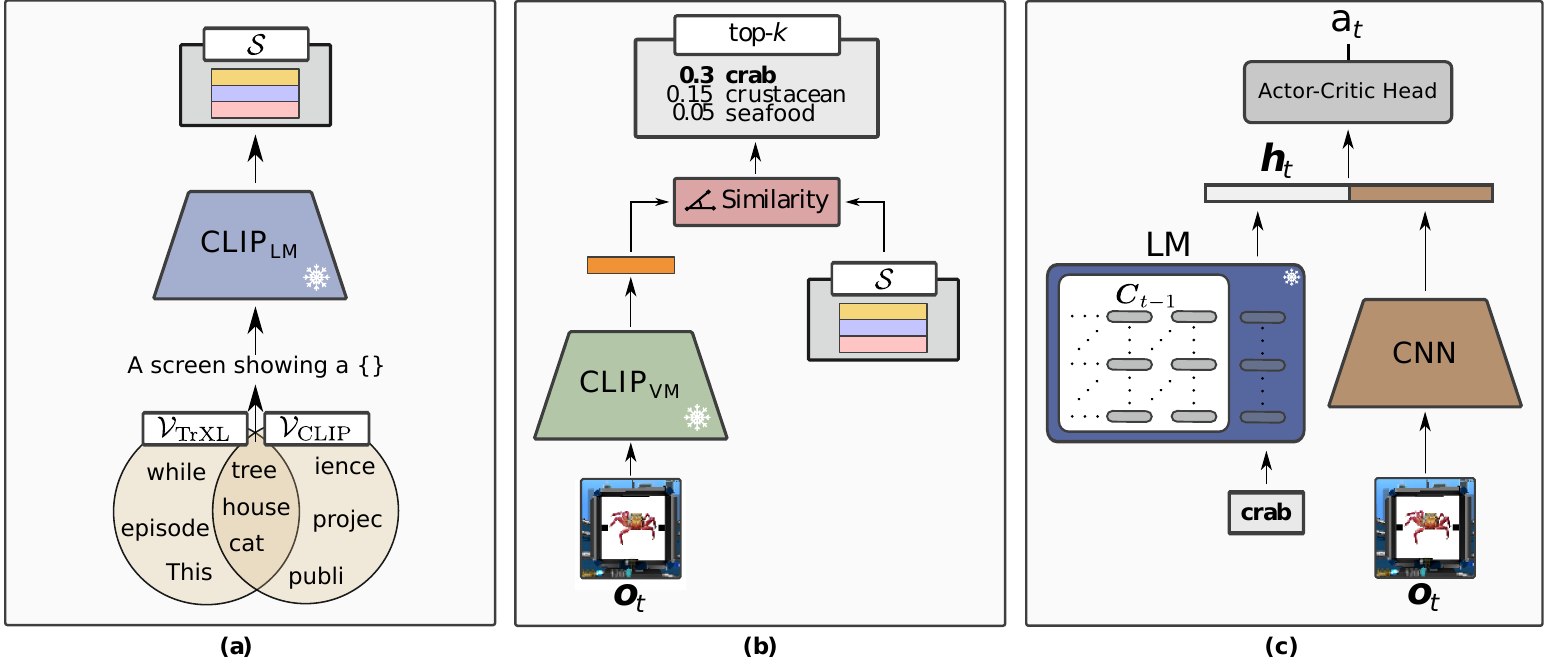}}
\caption{Architecture of SHELM. We compile a semantic database $\mathcal{S}$ by encoding prompt-augmented tokens from the overlapping vocabularies of CLIP and the LM \textbf{(a)}. Given an observation $\Bo_t$ we retrieve the top-$k$ embeddings present in $\mathcal{S}$ and select their corresponding text tokens \textbf{(b)}. These tokens are passed to the LM which represents the memory module of SHELM \textbf{(c)}. $\BC_{t-1}$ represents the memory cache of the LM which tracks past tokens.}
\label{fig:methodology}
\end{figure}

\subsection{Background}

HELM \citep{paischer_history_2022} was proposed as a framework for RL in partially observable environments.
It utilizes a pretrained language model (LM) as memory mechanism that compresses past observations. 
To map environment observations $\Bo_t \in \dR^n$ to the LM, it introduces the FrozenHopfield (FH) mechanism, which performs a randomized attention over pretrained token embeddings $\BE = (\Be_1, \dots, \Be_k)^\top \in \dR^{k \times m}$ of the LM, where $k$ is the vocabulary size and $m$ is the embedding dimension. 
Let $\BP \in \dR^{m \times n}$ be a random matrix with entries sampled independently from $\cN(0, n / m)$.
The FH mechanism performs
\begin{equation}
    \Bx_t = \BE^\top \operatorname{softmax} ( \beta \BE \BP \Bo_t), \label{eqn:frozen_hopfield}
\end{equation}
where $\beta$ is a scaling factor that controls the dispersion of $\Bx_t$ within the convex hull of the token embeddings. 
This corresponds to a spatial compression of observations to a mixture of tokens in the LM embedding space.
Since $\BP$ is random, the association of observations $\Bo_t$ with token embeddings $\Be_i$ is arbitrary, i.e., not meaningful. 
That is, the FH mechanism does not preserve semantics. 
For temporal compression, HELM leverages a pretrained LM.
At time $t$, HELM obtains a compressed history representation by
\begin{equation}
    \Bh_t = \operatorname{LM}(\Bc_{t-1}, \Bx_t)
\end{equation}
where $\Bc_t$ represents the context cached in the memory register of the LM up to timestep $t$.

More recent work has shown that the FH mechanism is prone to representation collapse if observations are visually similar to each other \citep{paischer_foundation_2022,paischer_toward_2022}.
They propose a new method, namely HELMv2, which substitutes the random mapping with a pretrained CLIP encoder.
Subsequently, they adopt a batch normalization layer \citep{ioffe_batch_2015} with fixed shifting and scaling parameters to transfer the image embedding to the language space.
Consequently, HELMv2 computes the inputs to the LM as
\begin{equation} \label{eqn:helmv2}
    \Bx_t = \operatorname{BN}_{\Bbe=\Bmu_E, \Bga=\Bsi_E}(\operatorname{CLIP}_{\text{VM}}(\Bo_t)),
\end{equation}
where $\operatorname{CLIP}_{\text{VM}}$ denotes the CLIP vision model and $\Bmu_E$ and $\Bsi_E$ denote mean and standard deviation of the embedded vocabulary $\BE$. 
This effectively fits the statistics of the image embeddings to those of the LM embeddings. 
Since the embedding spaces of CLIP and the LM were trained independently they are not semantically aligned. 
Therefore, also HELMv2 fails to preserve semantics of observations and, consequently, the memory mechanism of HELMv2 is not human-readable.

\subsection{Semantic HELM}\label{subsec:shelm}
Semantic HELM (SHELM) inherits the high-level architecture from HELM but introduces some changes to the memory module.
Similarly to HELMv2, we also use CLIP to embed environment observations.
However, we replace the batch normalization layer of HELMv2 with a token-retrieval mechanism.
We retrieve tokens that are similar to an observation in CLIP space and pass them to the LM in the form of text so they can be regarded as textual descriptions of environment observations. 

In a first step, we determine the overlap of the token vocabularies of CLIP and the LM. 
This is necessary to (i) to control for the number of tokens the LM receives per observation, and (ii) to avoid loss of information due to different tokenizers used by CLIP and the LM.
Thereby, we obtain a set of tokens $\cV$.
Since CLIP was pretrained on image-caption pairs, we augment each token $v \in \cV$ with a set of pre-defined prompts $\cP = \{p_1, p_2, \dots\}$\footnote{Following the original implementation at \url{https://github.com/openai/CLIP/blob/main/notebooks/Prompt_Engineering_for_ImageNet.ipynb}}.
The contents of $\cP$ are hyperparameters of our method and can be designed to specifically target certain aspects of the different environments. 
We embed a token $v$ in the CLIP output space by computing the average embedding of its prompt augmentations. 
That is, we define the function
\begin{equation}
    \operatorname{embed}(v) = \frac{1}{|\cP|} \sum_{p \in \cP} \operatorname{CLIP}_{\text{LM}}(\operatorname{concat}(p, v)).
\end{equation}
We do this for every $v \in \cV$, which results in a set $\cS$ of CLIP-embedded tokens
\begin{equation}
    \cS = \{\operatorname{embed}(v)\}_{v \in \cV}.
\end{equation} 

We denote by $\max^k$ an extension of the $\max$ operator that returns the subset of the $k$ largest elements in a set. 
For each observation we retrieve the $k$ most similar tokens in terms of cosine similarity by

\begin{equation}
    \cS^\ast = \max^k_{\Bs \in \cS} \operatorname{cossim}(\Bs, \operatorname{CLIP}_{\text{VM}}(\Bo_t)),
\end{equation}
where 
\begin{equation}
    \operatorname{cossim}(\Bx, \By) = \frac{\Bx^\top \By}{\|\Bx\| \|\By\|}.
\end{equation}

Note that $k = |\cS^\ast|$ is another hyperparameter of our method, namely the number of tokens that represent an observation.
Effectively, $k$ controls the degree of compression in the memory.

Finally, we embed single tokens $\Bv$ corresponding to the set of tokens in $\cS^\ast$ in the LM embedding space and pass them to the LM.
In this manner, we provide the LM with a textual representation of the current observation.
While HELM and HELMv2 also leverage the latent structure of a pre-trained LM, they do not explicitly represent the observations in the form of text, thus, do not provide a human-readable form of past observations.
Another improvement over HELMv2 is that SHELM removes the restriction of HELMv2 that the embedding spaces of CLIP and LM must have the same dimensionality.
In turn, any CLIP-like encoder can be used as semantic extractor for SHELM. 
\cref{fig:methodology} shows an illustration of the methodology of SHELM.

\section{Experiments}
\label{sec:exp_results}

First, we investigate in \cref{subsec:extract_semantics} whether CLIP can extract semantics of artificial scenes.
Next, we train SHELM on four different environments, namely MiniGrid \citep{chevalier-boisvert_minimalistic_2018}, MiniWorld \citep{chevalier-boisvert_miniworld_2018}, Avalon \citep{albrecht_avalon_2022}, and Psychlab \citep{leibo_psychlab_2018}.
We compare SHELM to HELMv2, LSTM (a recurrent baseline based on the LSTM architecture), and the popular Dreamerv2 \citep{hafner_mastering_2021} and Dreamerv3 \citep{hafner_dreamerv3_2023}.
We show that a semantic memory boosts performance in environments that are both heavily dependent on memory and photorealistic.
Finally, in \cref{subsec:ablations} we perform ablation studies on the benefit of semantics and the trade-off between interpretability and performance.

We train all HELM variants and LSTM with Proximal Policy Optimization (PPO,\citealp{schulman_proximal_2017}) on RGB observations.
Following \citet{paischer_history_2022}, we instantiate the LM with a pretrained TransformerXL (TrXL, \citealp{dai_transformer-xl_2019}) model. 
For training Dreamerv2 and Dreamerv3, we use the respective codebases\footnote{Dreamerv2: \url{https://github.com/danijar/dreamerv2}}\footnote{Dreamerv3: \url{https://github.com/danijar/dreamerv3}} and train on RGB observations.
We report results via IQM \citep{agarwal_deep_2021} and 95\% bootstrapped confidence intervals (CIs) unless mentioned otherwise.
We follow \citep{colas_hitchhikers_2019} and perform a Welch's t-test with a reduced significance level of $\alpha = 0.025$ at the end of training to test for statistical significance.
We elaborate on the architectural design and hyperparameter sweeps in \cref{sec:hyperparams}.

\subsection{Extracting semantics of virtual scenes}
\label{subsec:extract_semantics}

First, we analyse whether CLIP vision encoders are able to extract semantics from artificial scenes that are typically encountered in RL environments.
We compare the two smallest ResNet \citep{he_deep_2016} and ViT \citep{dosovitskiy_image_2021} architectures, namely RN50 and ViT-B/16, since those add the least amount of computational overhead to SHELM.
In this regard, we look at observations sampled via a random policy, provide them to the CLIP vision encoder and retrieve the closest tokens in the CLIP embedding space as explained in \cref{subsec:shelm}.
We find that the retrieval of tokens strongly varies between vision encoder architectures for MiniWorld and Avalon (see \cref{fig:clip_backbone_miniworld} and \cref{fig:clip_backbone_avalon} in the appendix).
The differences are especially prominent for MiniWorld environments, where the ViT-B/16 encoder recognizes shapes and colors, whereas RN50 ranks entirely unrelated concepts highest.
More photorealistic observations from the Avalon benchmark show a similar, but less pronounced trend.
We also observe a strong bias toward abstract tokens such as \emph{screenshot}, \emph{biome}, or \emph{render}.
We alleviate the retrieval of these tokens by including them in our prompts for retrieval.
Therefore, instead of using the prompt ``An image of a  $\texttt{<tok>}$'', we consider prompts such as ``A screenshot of a $\texttt{<tok>}$'', or ``A biome containing a $\texttt{<tok>}$'', where $\texttt{<tok>}$ stands for a token in $\cV$. 
We only use prompts for Avalon and Psychlab, since the effect of this prompting scheme was negligible on the other environments.
\cref{tab:env_prompts} in the appendix features the full list of prompts for the two environments.
We also add an analysis on using off-the-shelf captioning engines, such as BLIP-2 \citep{li_blip2_2023} in \cref{app:token_retrieval}.
We find that, while BLIP-2 correctly recognizes shapes, colors, and objects, it lacks accuracy in their compositionality.
Based on this analysis we use the ViT-B/16 encoder in combination with our designed prompts as semantics extractor for SHELM.

\begin{figure}[t]
    \centerline{\includegraphics[width=\textwidth]{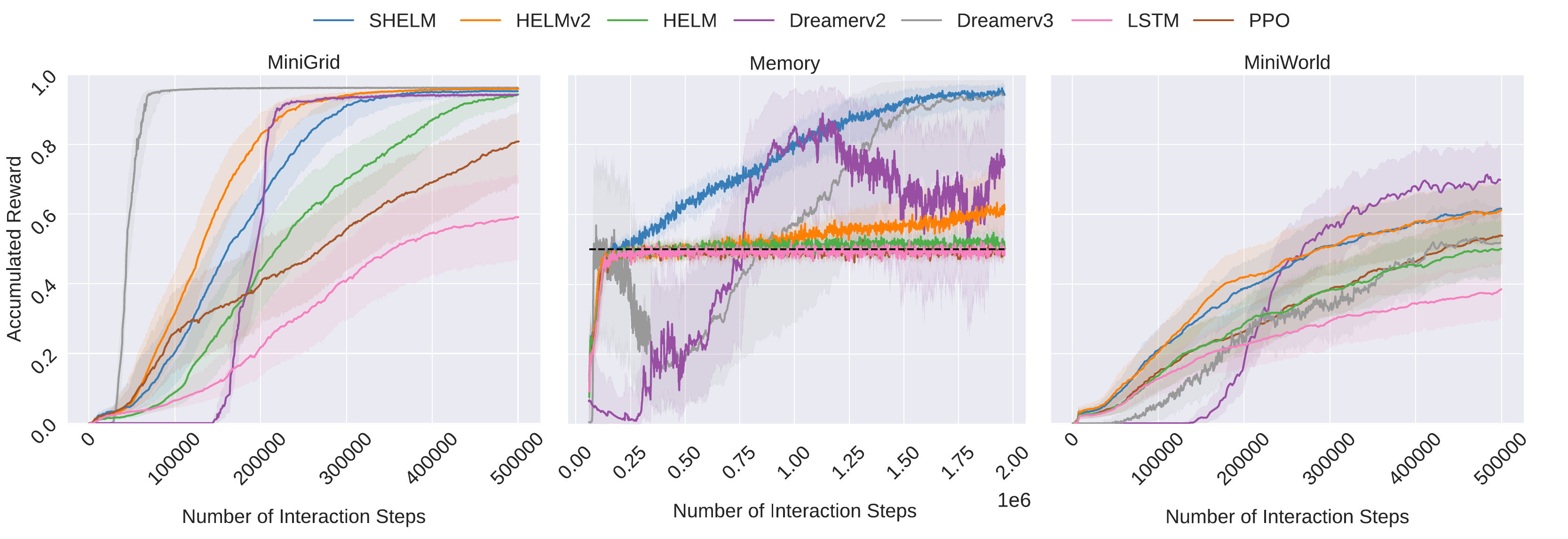}}
    \caption{\textbf{Left:} Accumulated reward for different methods on six MiniGrid environments, \textbf{Middle:} on the MiniGrid-Memory task, \textbf{Right:} on eight MiniWorld tasks. We report IQM and 95\% CIs across 30 seeds for each method.}
    \label{fig:grid_world_results}
\vskip -0.2in
\end{figure}

\subsection{MiniGrid}
\label{subsec:minigrid}

We compare all methods on a set of six partially observable grid-world environments as in as \citep{paischer_history_2022}.
Additionally, we train on the MiniGrid-MemoryS11-v0 environment, which we refer to as Memory.
The Memory task requires the agent to match the object in the starting room to one of the two objects at the end of a corridor.
The agent's view is restricted so it cannot observe both objects at the same time. 
Therefore, it needs to memorize the object in the starting room.
If the agent chooses the wrong object at the end of the corridor, it receives no reward and the episode ends.

\cref{fig:grid_world_results} (left and middle, respectively) shows the results across the six MiniGrid environments and the Memory environment.
On the MiniGrid environments, Dreamerv3 excels and converges much faster than any other method.
However, final performance is approximately equal for Dreamerv3, HELMv2, and SHELM.
Interestingly, although all MiniGrid environments are partially observable, they are solvable with a memory-less policy.
In the Memory environment, SHELM performs on par with Dreamerv3 but exhibits faster convergence and more stable training.
\cref{fig:mem_episode} visualizes the tokens passed to the memory of SHELM after sampling episodes from a trained policy.
SHELM primarily maps the ball to the token \emph{miner}, and the key to the token \emph{narrow}.
Although the retrieved tokens do not represent the semantically correct objects in the observations, they are still stored as two different language tokens in the memory.
This enables SHELM to discriminate between them and enables faster learning as mirrored in the learning curves.
The LSTM baseline suffers from poor sample efficiency and does not learn to utilize its memory within the budget of 2\,M interaction steps.
LSTM requires approximately 5\,M steps to solve the task.

\subsection{MiniWorld}
\label{subsec:miniworld}

The MiniWorld benchmark suite provides a set of minimalistic 3D environments.
In line with \citet{paischer_foundation_2022}, we select eight different tasks from MiniWorld and train our method and all baselines on those.
The tasks comprise collecting and placing objects, finding objects, or navigating through mazes.
A detailed description of each task can be found in \cref{app:envs}.
Prior work on the MiniWorld benchmark mostly used single environments to evaluate specialized approaches \citep{zheran_decoupling_2021,venuto_avoidance_2019,zha_rank_2021,khetarpal_options_2020}, or handcraft novel environments \citep{sahni_addressing_2019,hutsebaut_pretrained_2020}.
To the best of our knowledge, HELMv2 is the only method that was evaluated on all eight tasks and has been state of the art so far \citep{paischer_foundation_2022}.

\cref{fig:grid_world_results} (right) shows the results for all methods.
Interestingly, SHELM reaches performance on par with HELMv2, even though semantics can be extracted from the 3D observations to a certain degree (as explained in \cref{subsec:extract_semantics}).
Further, Dreamerv2 outperforms both and reaches state-of-the-art performance.
Surprisingly, Dreamerv3 attains a significantly lower score and performs on par with HELM and the memory-less PPO.
This might be due to suboptimal hyperparameters, even though \citet{hafner_dreamerv3_2023} claim that the choice of hyperparameters transfers well across environments.
LSTM again suffers from poor sample efficiency reaching the lowest scores out of all compared methods.
Finally, we again observe that PPO can in principle solve all tasks, which yields further evidence that partial observability does not automatically imply necessity for memory. 
Our results suggest that a memory mechanism can result in enhanced sample efficiency (SHELM vs PPO). 
However, memory is not imperative to solve these tasks.

\begin{figure}[t]
    \begin{minipage}{.4\columnwidth}
    \resizebox{\textwidth}{!}{ 
    \begin{tabular}{cc}
        \toprule
        Method & Mean Human Normalized Score \\
        \midrule
        HELM & 0.143 $\pm$ 0.010 \\
        HELMv2 & 0.145 $\pm$ 0.010\\
        PPO & 0.146 $\pm$ 0.010 \\
        SHELM & 0.155 $\pm$ 0.011 \\
        Dreamerv2 & 0.168 $\pm$ 0.012\\
        \bottomrule
    \end{tabular}
    }
    \end{minipage}
    \begin{minipage}{0.59\columnwidth}
         \centerline{\includegraphics[width=0.85\textwidth]{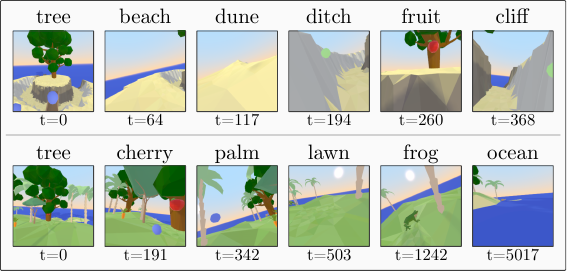}}
    \end{minipage}
    \caption{\textbf{Left:} Mean and standard deviation of human normalized score across all tasks on the Avalon test worlds after 10\,M timesteps. Boldface indicates maximum average score. \textbf{Right:} Two sample episodes and their corresponding token mappings for episodes sampled from Avalon with a random policy.}
    \label{fig:avalon}
    \vskip -0.1in
\end{figure}

\subsection{Avalon}
\label{subsec:avalon}

Avalon is an open-ended 3D survival environment consisting of 16 different tasks.
The aim for the agent is to survive as long as possible by defending against predators, hunting animals, and eating food in order to restore energy.
An episode ends if the agent has no energy left.
This can happen if the agent receives environmental damage (e.g. from falling), is being killed by a predator, or does not eat frequently.
The agent receives a dense reward as the difference in energy between consecutive timesteps.
Additionally, it receives a delayed reward upon successful completion of a task, e.g., eating food.
The observation space are RGBD images as well as proprioceptive input that comprise, e.g., the agent's energy.
We adopt the same architecture and training strategy as \citet{albrecht_avalon_2022} for all HELM variants.
Specifically, we add the history compression branch to the baseline PPO agent and train on all 16 tasks including their difficulty curriculum.
The history compression branch is only based on RGB images and does not receive the proprioceptive state.
We also train Dreamerv2, and a memory-less baseline (PPO), which is identical to the PPO baseline in \citet{albrecht_avalon_2022}.
Due to computational constraints and since we observed superior performance of Dreamerv2 on 3D environments, we neglect Dreamerv3 and LSTM and train on a limited budget of 10\,M interaction steps.
We elaborate on training details and hyperparameter selection in \cref{sec:hyperparams}.
The final performance of an agent is measured in terms of mean human normalized scores on a curated set of 1000 test worlds.

The results are shown in \cref{fig:avalon}, left. 
For detailed results per task see \cref{tab:avalon_hns_10m} in the appendix.
SHELM and Dreamerv2 yield the highest performance on average after 10\,M interaction steps.
However, the difference to the memory-less PPO is not statistically significant.
To further investigate this finding we train PPO for 50\,M interaction steps and compare it to the baseline results reported in \citet{albrecht_avalon_2022} in \cref{app:additional_results}.
Indeed, we find that our memory-less baseline attains scores on-par with memory-based approaches trained for 50\,M interaction steps.
This yields further evidence that Avalon does not necessitate the addition of a memory mechanism.
Finally, we can glance at SHELM's memory to identify failure cases.
We show observations of two episodes and their token correspondences for SHELM in \cref{fig:avalon}, right.
The observations are mostly mapped to semantically meaningful tokens that represent parts in the image observations.
However, we also observe some cases where CLIP retrieves tokens that are semantically related, but not correct, which we show in \cref{fig:avalon_episodes_appendix}.

\begin{figure}[t]
    \begin{minipage}{.5\columnwidth}
        \centerline{\includegraphics[width=.9\columnwidth]{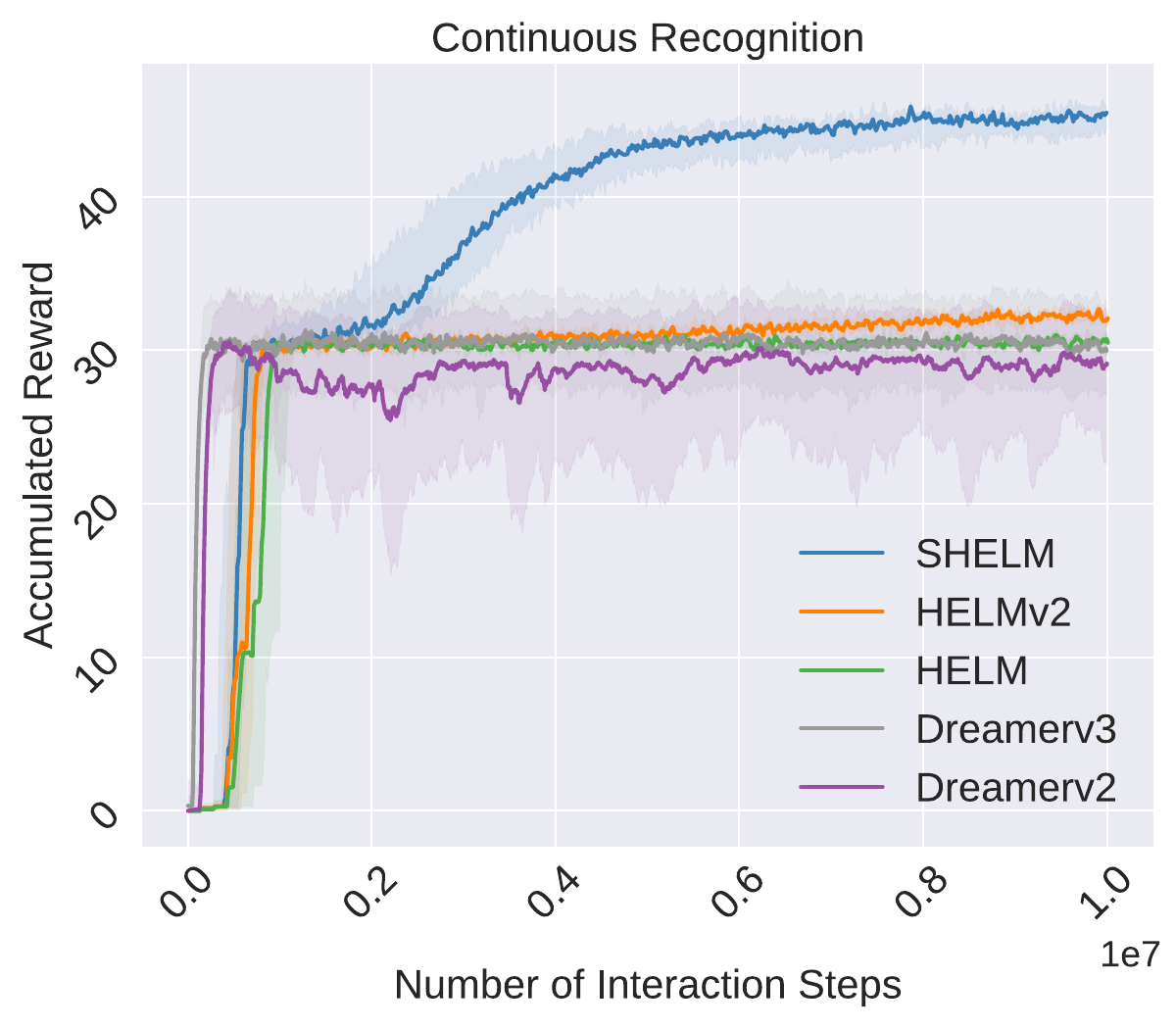}}       
    \end{minipage}
    \begin{minipage}{0.5\columnwidth}
         \centerline{\includegraphics[width=.9\columnwidth]{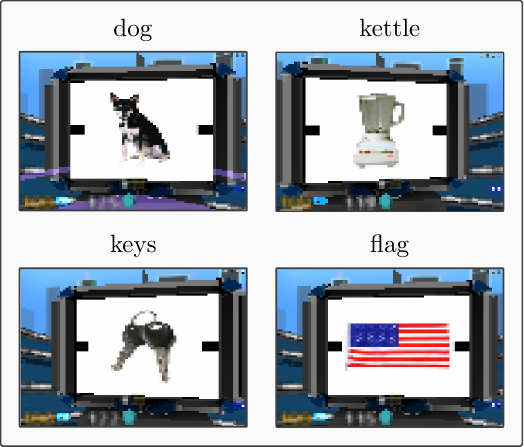}}    
    \end{minipage}
    \caption{Performance for all methods on CR task from Psychlab. We report IQM and 95\% CIs across 5 seeds (\textbf{left}). Observation and corresponding tokens for SHELM on CR environment (\textbf{right}).}
    \label{fig:psychlab}
    \vskip -0.1in
\end{figure}

\subsection{PsychLab}
\label{subsec:psychlab}

The Psychlab environment suite \citep{leibo_psychlab_2018} consists of 8 complex tasks and was designed to isolate various cognitive faculties of RL agents including vision and memory.
The continuous recognition (CR) task of Psychlab was specifically designed to target the memory capacity of an agent.
In this task the agent is placed in front of a monitor which displays a stream of objects.
The agent then needs to identify whether it has already seen a particular object by swiping either left or right.
If the agent correctly identifies an object it receives a reward of +1.
A policy that always swipes in one direction achieves a reward of around 30, which we refer to as the random baseline.
Every reward higher than that indicates that an agent actually utilizes its memory.
This task is ideal to evaluate the memory capacity of an agent, since episodes last for about 2000 timesteps and usually require remembering up to 50 different objects.

We train all methods for 10\,M interaction steps on the CR task (\cref{fig:psychlab}).
We neglect the memory-less baseline because this task is unsolvable without memory.
SHELM indeed learns to effectively utilize its memory within 10\,M interaction steps and significantly outperforms all competitors.
Other approaches require interaction steps in the range of billions until convergence \citep{parisotto_stabilizing_2020,fortunato_generalization_2019}.
These works report human normalized scores which are not publicly available, therefore we cannot compare SHELM to those.
HELMv2 is the only competitor that attains a performance better than random, but reaches significantly lower scores than SHELM.
Surprisingly, both variants of Dreamer do not exceed performance of the random baseline.
Finally, we inspect the memory of SHELM and show the token mappings that are passed on to the memory module for some sample observations in \cref{fig:psychlab}.
In most cases SHELM assigns semantically correct tokens to the displayed objects.
However, we also show some cases where the token retrieval of SHELM conflates different objects in \cref{fig:dmlab_samples}.
We find that this can mostly be attributed to the low resolution of observations.
SHELM can recover from these failure cases when using higher resolutions (see \cref{fig:dmlab_corrects}).

\subsection{Ablation studies}
\label{subsec:ablations}

\begin{wrapfigure}{r}{.5\textwidth}
    \centerline{\includegraphics[width=.5\textwidth]{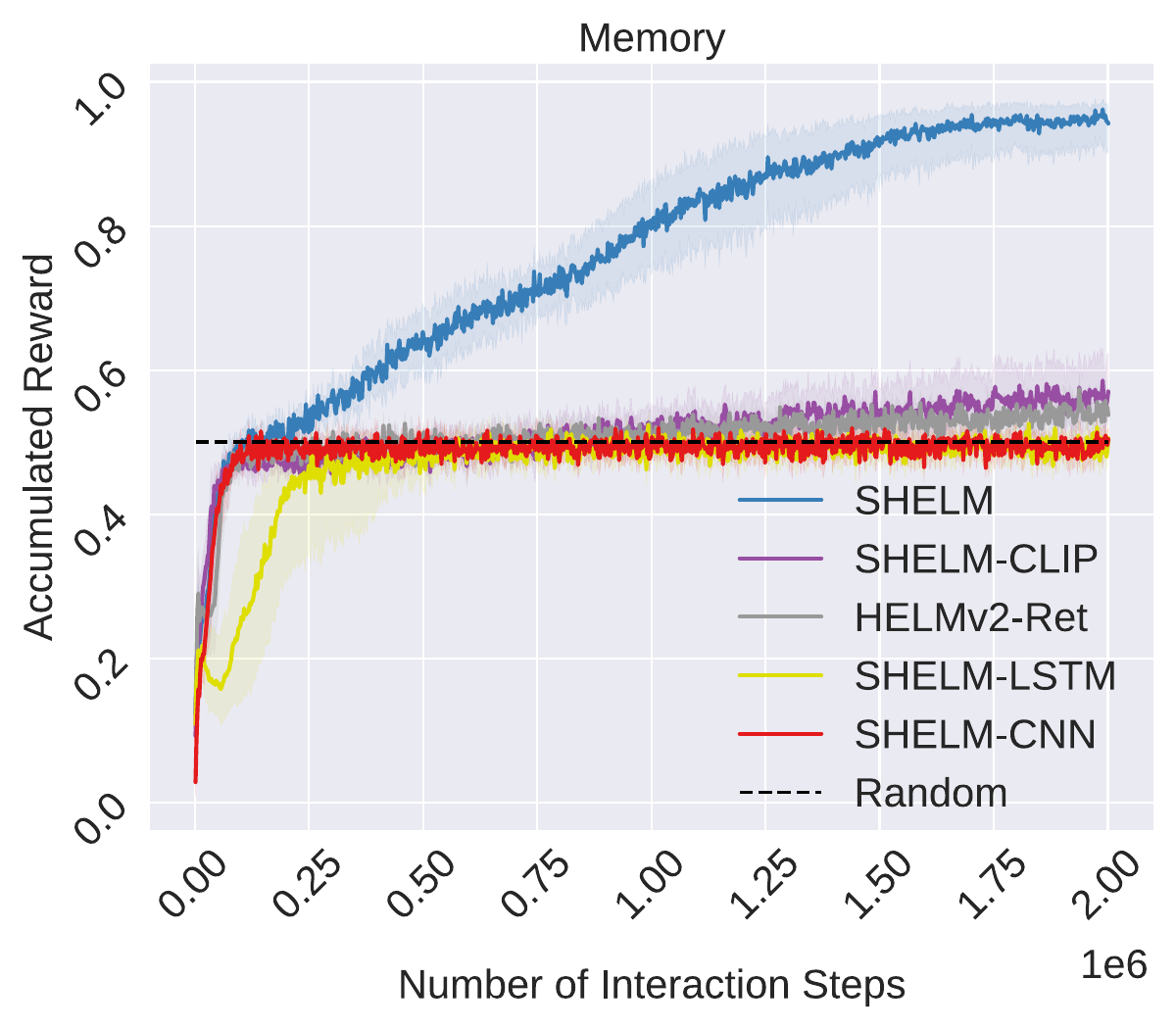}}
    \caption{Ablation study on the effect of semantics, the influence of task-specific features in the history and the current timestep, and the importance of the pretrained LM. We report IQM and 95\% CIs across 30 seeds on the MiniGrid-Memory task.}
    \label{fig:memory_ablations}
    \vskip -0.1in
\end{wrapfigure}

\paragraph{Are semantics important?}
We slightly alter the HELMv2 implementation from \citep{paischer_foundation_2022,paischer_toward_2022} by retrieving the closest tokens in the language space after their frozen batch normalization layer, and finally passing those to the LM.
This setting is similar to SHELM in that the LM receives tokens in the form of text. 
However, semantics are not preserved since visual features are merely shifted to the language space.
We call this setting HELMv2-Ret and train it on the Memory environment (see \cref{fig:memory_ablations}).
We find that if the mapping from observation to language is arbitrary, the performance decreases drastically.

\paragraph{Is it important to learn task-specific features?}
SHELM would be even more interpretable if not only the memory module but also the branch processing the current observation (CNN in \cref{fig:methodology}) could utilize language tokens.
Therefore, we substitute the CNN with a CLIP vision encoder (SHELM-CLIP).
An important consequence of this methodological change is that even features from the current observation can be interpreted by our retrieval mechanism.
However, \cref{fig:memory_ablations} suggests that it is vital to learn task-specific features. 

\paragraph{Should the history branch operate on task-specific features?}
To answer this question we substitute the CLIP encoder in the history branch with the CNN encoder learned at the current timestep (SHELM-CNN).
An immediate drawback of that is the loss of readability, since we do not represent the observations in text form anymore.
Further, we again observe a drastic drop in performance.
The reason for this is that the task-specific features must be learned before they actually provide any benefit to the history branch.
Thus, we prefer to keep abstract semantic features in text form since that does not require any training and has the additional benefit of being human-readable.

\paragraph{Is a pretrained language encoder required?}
We replace the LM with an LSTM operating on the text tokens (SHELM-LSTM).
This setting resulted in performance equivalent to randomly choosing an object at the end of the corridor.
However, we believe that after longer training SHELM-LSTM can eventually learn to solve the task, since the LSTM baseline also solved the task after longer trainng.
Thus, the simpler and more sample efficient method is to maintain the frozen pretrained encoder instead of learning a compressed history representation.

We show additional results for an ablation study where we exchange the ViT-B/16 vision encoder with a RN50 in \cref{app:ablations}.
SHELM appears to be quite sensitive to the selection of vision encoder, which corroborates our qualitative findings in \cref{subsec:extract_semantics}.

\section{Related work}

\paragraph{RL and partial observability}
RL with incomplete state information can necessitate a memory for storing the history of observations an agent encountered.
However, storing the entire history is often infeasible. 
History Compression tries to answer the question of what information to store in a stream of observations \citep{
schmidhuber_learning_1992,schmidhuber_continuous_1993,zenke_continual_2017,kirkpatrick_overcoming_2016,schwarz_progress_2018,ruvolo_ella_2013}.
A plethora of prior works have used history compression to tackle credit assignment \citep{arjona-medina_rudder_2019,patil_align-rudder_2022,widrich_modern_2021,holzleitner_convergence_2021}, and partial observability \citep{hausknecht_deep_2015,vinyals_grandmaster_2019,berner_dota_2019,pleines_generalization_2022}.
The memory maintained by an agent can either be external \citep{hill_grounded_2021,wayne_unsupervised_2018}, or directly integrated into the feature extraction pipeline via the network architecture.
An orthogonal approach for history compression is training recurrent dynamics models
\citep{ha_recurrent_2018,pasukonis_evaluating_2022,hafner_dream_2020,hafner_mastering_2021}.
We believe language is very well suited as medium for compression to summarize past events, as suggested by \citet{paischer_history_2022}.

\paragraph{Language in RL}
Language provides useful abstractions for RL.
These abstractions can be leveraged to represent high-level skills \citep{sharma_skill_2022,jiang_language_2019,jacob_multitasking_2021}.
LMs have been used to improve exploration in text-based environments \citep{yao_keep_2020}, or in visual environments via a language oracle \citep{mu_improving_2022}.
Pretrained vision-language models (VLMs) provide abstract embedding spaces that can be used for exploration in visual environments \citep{tam_semantic_2022}.
We leverage VLMs to spatially compress visual observations to language tokens.
Furthermore, pretrained LMs were leveraged for (i) initializing policies in text-based environments \citep{li_pre-trained_2022}, (ii) grounding to various environments \citep{andreas_learning_2018,huang_grounded_2022,carta_grounding_2022,hill_grounded_2021}, (iii) sequence modeling in the offline RL setup \citep{reid_can_2022}, and (iv) generating high-level plans \citep{huang_language_2022,huang_inner_2022,wang_describe_2023,singh_progprompt_2022,ichter_do_2022,liang_code_2022,dasgupta_collaborating_2023,du_guiding_2023,shah_lmnav_2022,ahn_do_2022,zeng_socratic_2022}.
Other works train agents to generate descriptions of virtual scenes \citep{yan_intra_agent_2022}, or thought processes of humans \citep{hu_thought_2023}.
Additionally, language has been used for augmenting the reward function \citep{wang_reinforced_2019,bahdanau_learning_2019,goyal_using_2019,carta_eager_2022,kwon_reward_2023}, or learning a dynamics model \citep{zhong_improving_2022,zhong_silg_2021,zhong_rtfm_2020,wu_read_2023}.
To manage a language-based memory, \citet{park_generative_2023} stores a record of an agent's memory which comprises different levels of abstraction.
Importantly, all agent-environment interactions in their work are scripted, while our agent enables a memory based on language for visual inputs.

\paragraph{Language for interpretability}
In the realm of supervised learning, a plethora of prior works had used language as human-interpretable medium to explain classification decisions in computer vision \citep{hendricks_generating_2016,hendricks_grounding_2018,park_multimodal_2018,zellers_from_2019,hernandez_natural_2022}, or in natural language processing \citep{andreas_analogs_2017,zaidan_modeling_2008,camburu_esnli_2018,rajani_explain_2019,narang_wt5_2020}.
Interpretability methods in RL are scarce and usually follow a post-hoc approach \citep{puiutta_xrl_2020}.
Intrinsically interpretable models are designed to be inherently interpretable even during training time and are preferable over post-hoc approaches \citep{rudin_interpretableml_2021}.
They often restrict the complexity of the model class, which in turn results in reduced performance of the agent. 
Therefore, \citep{glanois_survey_2021} propose to adopt a modular approach to interpretability. 
To this end, our work focuses on the memory module. 
This enables us to provide some extent of intrinsic interpretability while exceeding performance of existing (non-interpretable) methods on tasks that necessitate memory.

\paragraph{Foundation models}
The advent of the Transformer architecture \citep{vaswani_attention_2017} gave rise to foundation models (FMs, \citealp{bommasani_opportunities_2021}), such as GPT-3 \citep{brown_language_2020}.
As shown by \citet{petroni_language_2019,talmor_olmpics_2020, kassner_are_2020,mahowald_dissociating_2023}, pretrained LMs can learn abstract symbolic rules and show sparks of reasoning.
We leverage their abstraction capabilities for history compression in RL.
Further, vision FMs have been demonstrated to be well adaptable to foreign domains \citep{adler_cross-domain_2020,evci_head2toe_2022,ostapenko_foundational_2022,parisi_unsurprising_2022}.
Our approach combines language-based FMs with vision-langauge models, such as CLIP \citep{radford_learning_2021} or ALIGN \citep{jia_scaling_2021}.
We use CLIP to obtain language tokens that semantically correspond to concepts present in synthetic environments.

\section{Limitations}
\label{sec:limitations}

\paragraph{Token-level abstraction}
One potential shortcoming of our method is that our retrieval is based on single tokens.
However, we have shown that for environments where one or a few objects are important, this is sufficient.
Further, our approach is very flexible and the semantic database can easily be augmented with more detailed descriptions of objects and their relationships.
We aim to investigate more in this direction in future work.

\paragraph{Wall clock time} 
A current limitation of SHELM is wall clock time.
The rollout phase is particularly expensive since each timestep needs to be propagated through the LM.
Despite that, it is still more efficient than, e.g., Dreamerv2.
This is due to the fact that the memory mechanism is kept frozen and need not be updated durint the training phase.
A potential solution for decreasing the complexity of the rollout phase would be distillation of the LM into smaller actor networks as in \citep{parisotto_efficient_2021}.

\paragraph{Modality gap}
Our semantic mapping is limited by the inherent ability of CLIP vision encoders to extract semantically meaningful features in synthetic environments.
However, CLIP suffers from the modality gap \citep{liang_mind_2022}, i.e., a mis-alignment between image and text embedding spaces.
In the future, we aim at incorporating methods that mitigate the modality gap \citep{furst_cloob_2022,ouali_black_2023}.

\paragraph{Distribution shift}
We use a pretrained CLIP model to retrieve tokens that are similar to visual observations.
However, CLIP was pretrained on large-scale data crawled from the web.
We have shown that it can still extract semantics of synthetic environments when those are sufficiently photorealistic, i.e., MiniWorld, Avalon, or Psychlab.
Further, prompting can enhance the quality of the retrieval.
For environments such as MiniGrid, the retrieval yields tokens that do not correspond to aspects in the image.
However, \citet{gupta_foundation_2022} has shown that CLIP can handle task-specific texts for MiniGrid, thus, we would like to investigate the effect of augmenting our semantic database with such texts in the future.

\section{Conclusion}
\label{sec:conclusion}
In many real-world scenarios an agent requires a memory mechanism to deal with partial observability.
Current memory mechanisms in RL act as a black box where it is not comprehensible for humans what pieces of information were stored.
To solve this problem, we proposed a new method called Semanic HELM that represents past events in form of human-readable language by leveraging pretrained vision-language models.
We showed compelling evidence that even for synthetic environments our memory mechanism can extract semantics from visual observations.
Further, SHELM outperforms strong baselines on photorealistic memory-dependent environments through its human-readable memory module.
In cases where SHELM fails to extract semantics from observations, we can investigate the cause by inspection of the memory module.
Even in such cases, SHELM mostly performs on-par with other memory-based approaches.

We believe that we can further enhance our method by generating full captions from history observations instead of only a small number of words.
This could enable (i) a long-term memory that textually summarizes all captions of corresponding observations, (ii) a potential for planning in form of text from visual observations similar to \citet{patel_pretrained_2023}, and (iii) modeling dynamics of an environment in language space.

\section*{Acknowledgements}
The ELLIS Unit Linz, the LIT AI Lab, the Institute for Machine Learning, are supported by the Federal State Upper Austria. We thank the projects AI-MOTION (LIT-2018-6-YOU-212), DeepFlood (LIT-2019-8-YOU-213), Medical Cognitive Computing Center (MC3), INCONTROL-RL (FFG-881064), PRIMAL (FFG-873979), S3AI (FFG-872172), DL for GranularFlow (FFG-871302), EPILEPSIA (FFG-892171), AIRI FG 9-N (FWF-36284, FWF-36235), AI4GreenHeatingGrids(FFG- 899943), INTEGRATE (FFG-892418), ELISE (H2020-ICT-2019-3 ID: 951847), Stars4Waters (HORIZON-CL6-2021-CLIMATE-01-01). We thank Audi.JKU Deep Learning Center, TGW LOGISTICS GROUP GMBH, Silicon Austria Labs (SAL), University SAL Labs initiative, FILL Gesellschaft mbH, Anyline GmbH, Google, ZF Friedrichshafen AG, Robert Bosch GmbH, UCB Biopharma SRL, Merck Healthcare KGaA, Verbund AG, GLS (Univ. Waterloo) Software Competence Center Hagenberg GmbH, T\"{U}V Austria, Frauscher Sensonic, TRUMPF and the NVIDIA Corporation.

\bibliography{neurips2023}
\bibliographystyle{apalike}

\newpage
\appendix

\section{Environments}
\label{app:envs}
We choose 8 diverse 3D environments of the MiniWorld benchmark suite:

\begin{itemize}
    \item \textbf{CollectHealth:} The agent spawns in a room filled with acid and must collect medikits in order to survive as long as possible.
    \item \textbf{FourRooms:} The agent must reach a red box that is located in one of four interconnected rooms.
    \item \textbf{MazeS3Fast:} A procedurally generated maze in which the agent needs to find a goal object.
    \item \textbf{PickupObjs:} Several objects are placed in a large room and must be collected by the agent.
    Since the agent receives a reward of 1 for each collected object, the reward is unbounded.
    \item \textbf{PutNext:} Several boxes of various colors and sizes are placed in a big room. The agent must put a red box next to a yellow one.
    \item \textbf{Sign:} The agent spawns in a U-shaped maze containing various objects of different colors. One side of the maze contains a sign which displays a color in written form. The aim is to collect all objects in the corresponding color.
    \item \textbf{TMaze:} The agent must navigate towards an object that is randomly placed at either end of a T-junction. 
    \item \textbf{YMaze:} Same as TMaze, but with a Y-junction.
\end{itemize}

We neglect the OneRoom and the Hallway environments, since those are easily solved by all compared methods.
Further, we neglect the Sidewalk environment since it is essentially the same task as Hallway with a different background.
Since the rewards of PickupObjs and CollectHealth are unbounded, we normalize them to be in the range of $[0,1]$, which is the reward received in all other environments.
For a more detailed description of the MiniGrid environments we refer the reader to \citet{paischer_history_2022}.

\section{Token retrieval for synthetic environments}
\label{app:token_retrieval}

\begin{figure}
    \centering
    \includegraphics[width=.75\textwidth]{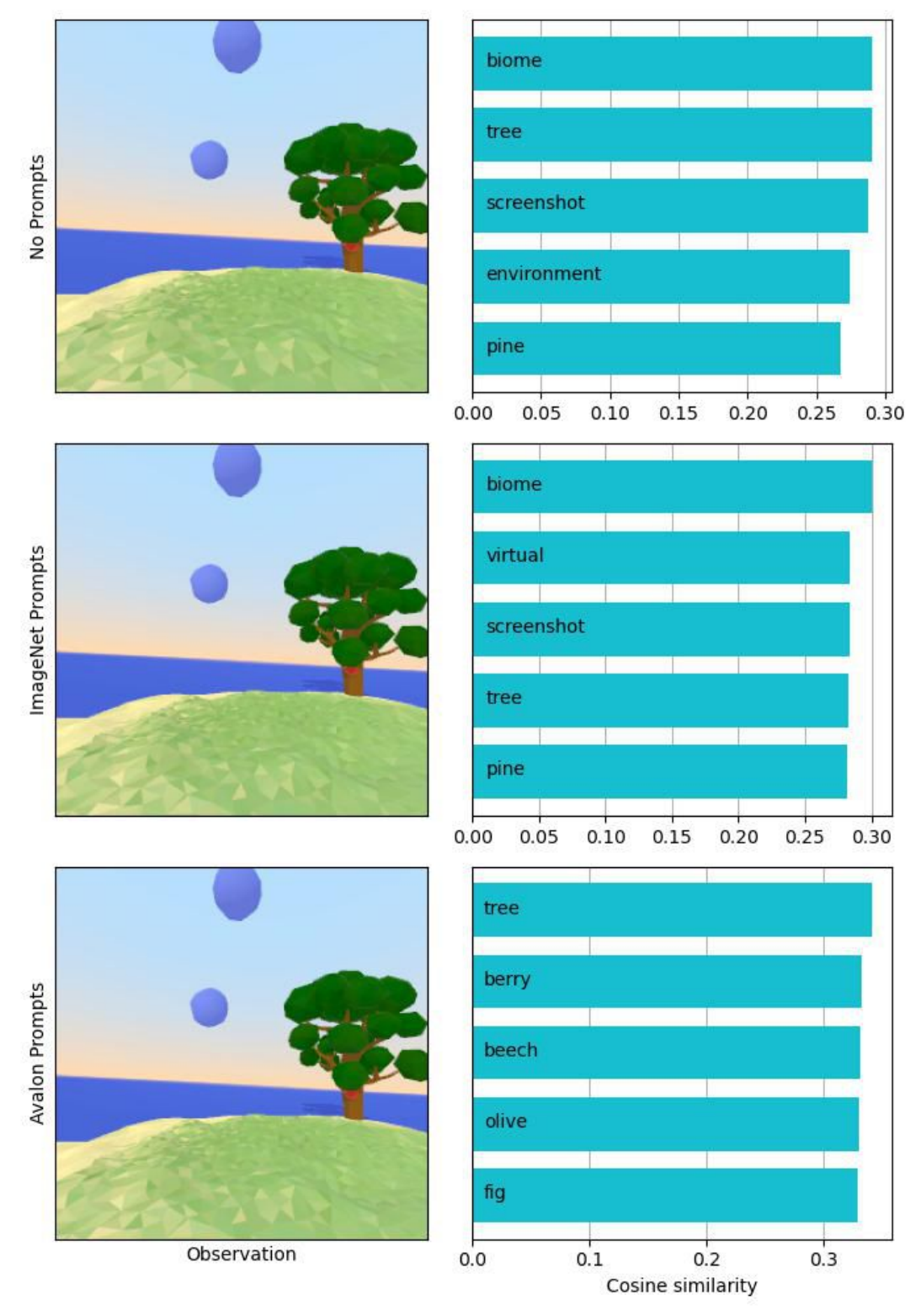}
    \caption{Token rankings for ViT-B/16 encoder of CLIP on Avalon observations. Tokens are encoded with the CLIP language encoder without prompt (\textbf{top}), with ImageNet specific prompts (\textbf{middle}), or prompts designed for Avalon (\textbf{bottom}).}
    \label{fig:prompts_retrieval_avalon}
\end{figure}

Since we perform retrieval on a token level we investigate the effect of augmenting tokens with different prompts, and the effect of different vision encoders on retrieval performance.
We analyse the former at the example of the Avalon environment.
\cref{fig:prompts_retrieval_avalon} shows some examples.
We observed that simply encoding single tokens in CLIP space results in a bias toward abstract tokens such as \emph{biome}, or \emph{screenshot}.
The same occurs for using the prompts originally used for zero-shot classification on the ImageNet dataset\footnote{available at https://github.com/openai/CLIP/tree/main} \citep{deng_imagenet_2009}.
However, one can alleviate this bias by including these frequently retrieved tokens in the prompt itself (see \cref{fig:prompts_retrieval_avalon}, bottom).
However, we found this strategy to be effective only for Avalon and Psychlab environments.
The sets of prompts for both environments can be found in \cref{tab:env_prompts}.
For MiniGrid and MiniWorld we retrieve single tokens without any prompt.

\begin{figure}
    \centering
    \includegraphics[width=\textwidth]{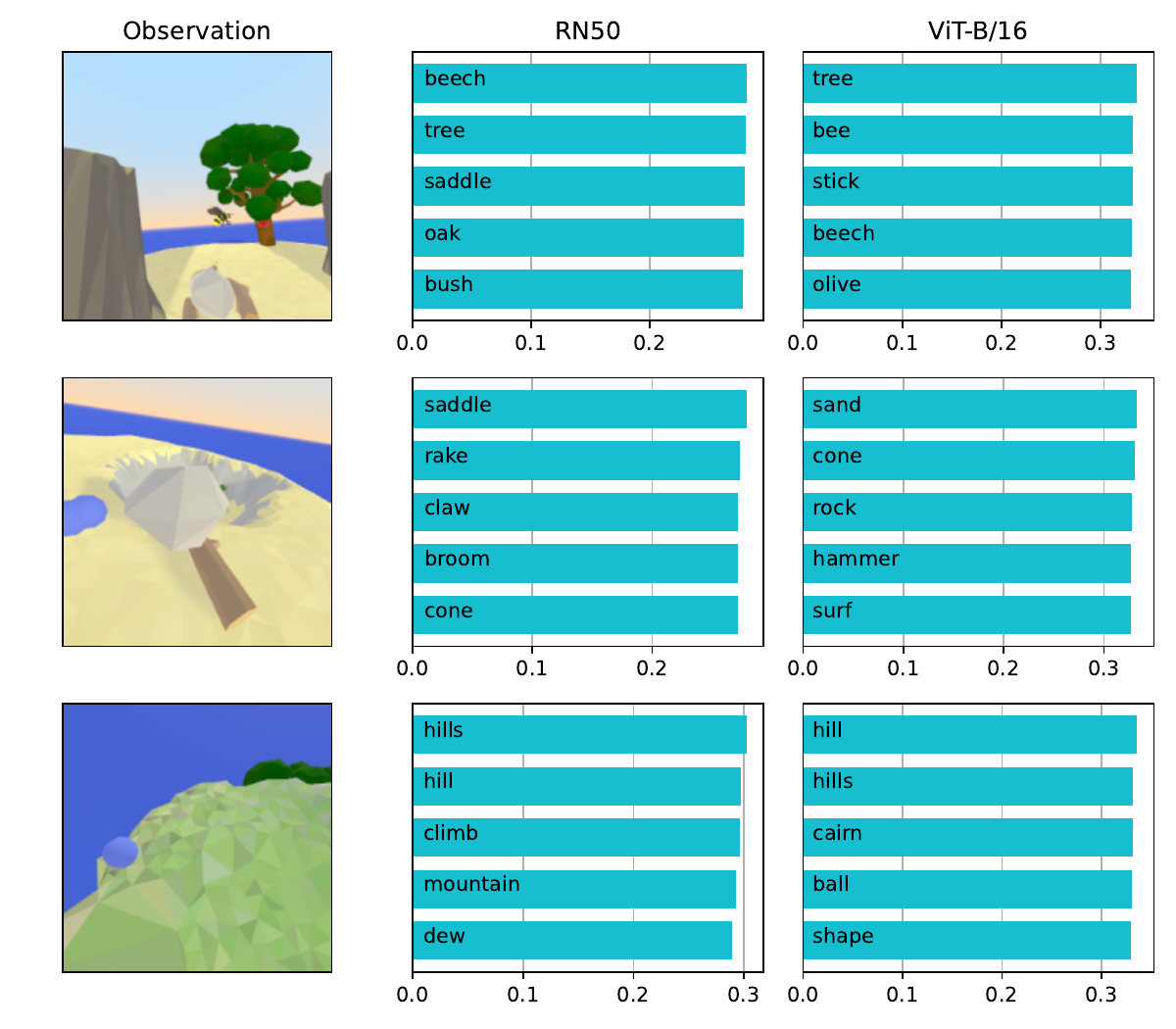}
    \caption{Token rankings for RN50 and ViT-B/16 encoders of CLIP on Avalon observations.}
    \label{fig:clip_backbone_avalon}
\end{figure}

\begin{figure}
    \centering
    \includegraphics[width=\textwidth]{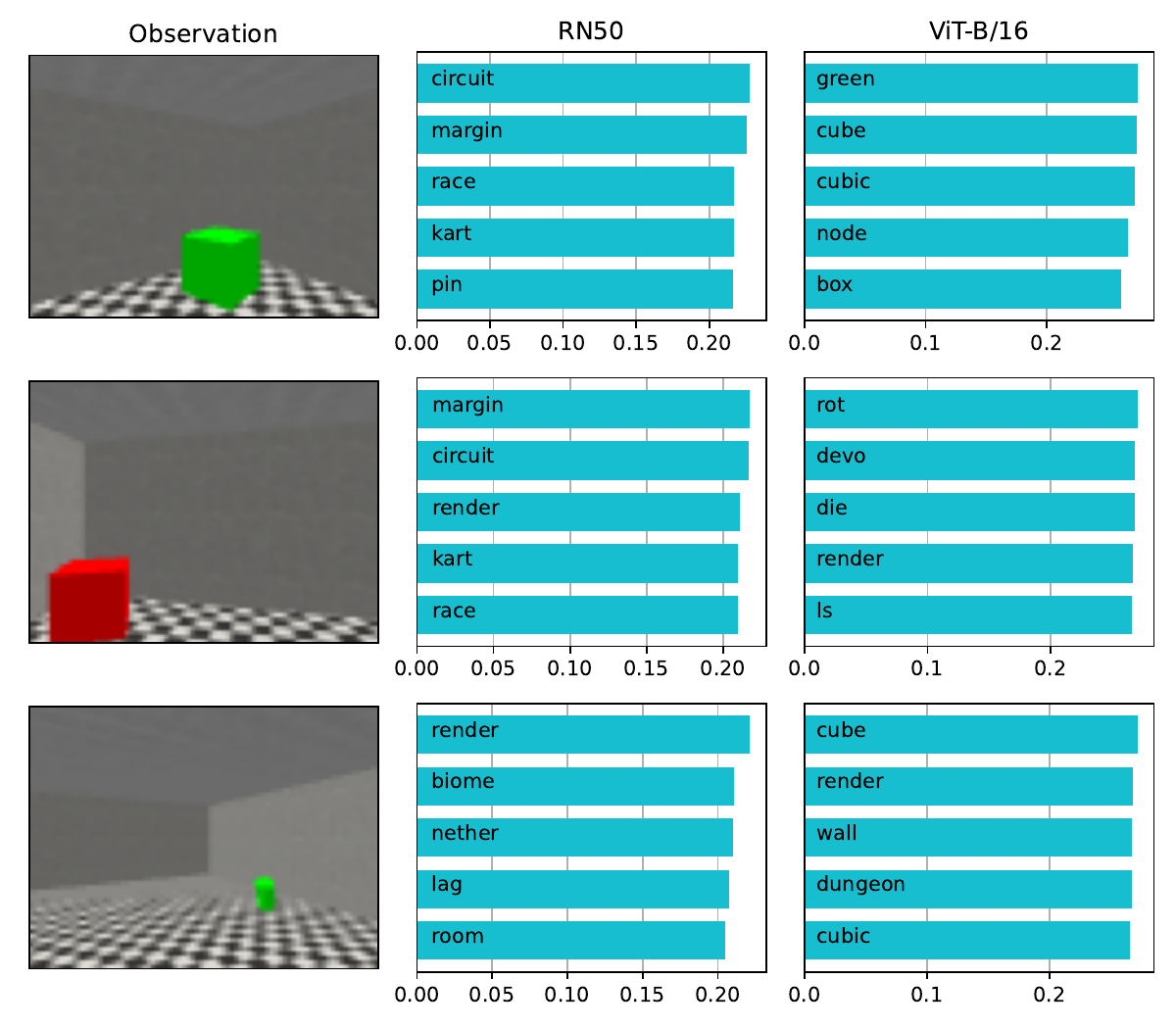}
    \caption{Token rankings for RN50 and ViT-B/16 encoders of CLIP on MiniWorld observations.}
    \label{fig:clip_backbone_miniworld}
\end{figure}

Next, we investigate the influence of the choice of vision encoder architecture, e.g., RN50 vs ViT-B/16.
We only consider those encoders since they induce the least complexity on our history compression pipeline.
We show the closest tokens in CLIP space for observations of MiniWorld (see \cref{fig:clip_backbone_miniworld}) and Avalon environments (see \cref{fig:clip_backbone_avalon}).
For MiniWorld, CLIP can extract shapes and colors.
However, the retrievals are very noisy, which is mirrored in the attained score of SHELM.
For Avalon, however, the token retrievals are more convincing.
Generally, we find that retrievals using the RN50 encoder of CLIP tend to be more noisy than retrieval utilizing a ViT-B/16 encoder.

\begin{figure}
    \centering
    \includegraphics[width=\textwidth]{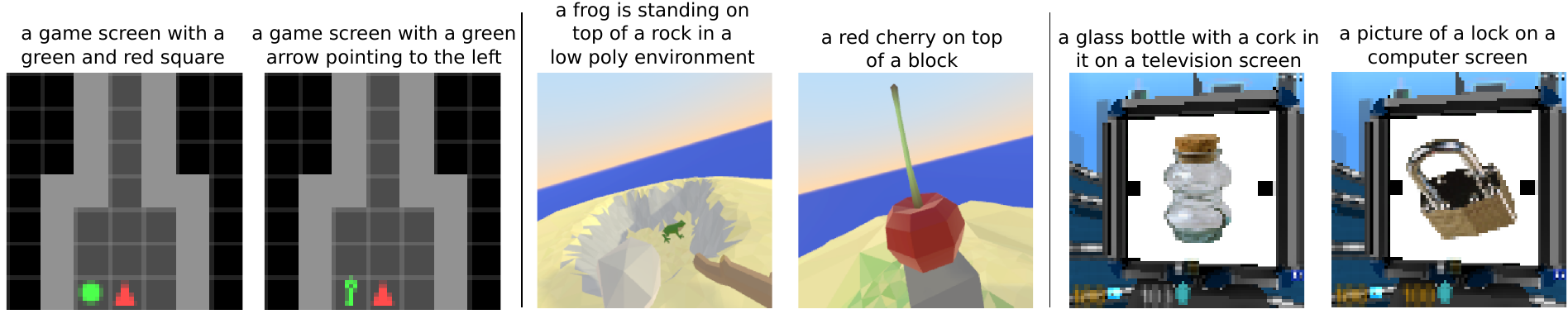}
    \caption{Captions generated via BLIP-2 for observations of the MiniGrid-Memory task, Avalon and Psychlab.}
    \label{fig:blip-2_captions}
\end{figure}

Instead of using CLIP and our token retrieval we may be able to use pretrained image captioning systems, such as BLIP-2 \citep{li_blip2_2023} directly.
To investigate this, we use BLIP-2 to generate captions for MiniGrid-Memory, Avalon and Psychlab (\cref{fig:blip-2_captions}).
The captions generated for the MiniGrid-Memory in the left column contain correctly detected colors (green and red), and shapes (square), but incorrectly combines them, e.g. there is neither a green square, nor a green arrow pointing to the left.
A similar trend can be observed for Avalon in the middle comlumn, where BLIP-2 recognizes a frog, but incorrectly assumes that it stands on top of a rock, while the rock is clearly beside the frog in the foreground.
The second observation for Avalon is very accurate though.
For the two Psychlab observations in the right column, BLIP-2 correctly describes the objects on the screen, but also mentions redundant information, such ''the objects are on a screen``.
Overall, BLIP-2 correctly recognizes shapes, colors, and different objects, but is often incorrect on their compositionality.
Further, a drawback of using off-the-shelf captioning engines is that we cannot control for the context length for each observation.
This can quickly result in excessive context lengths, which in turn, results in a drastic increase of computational complexity.

\begin{table}[h]
    \centering
    \caption{Prompts used for token retrieval for the Avalon and Psychlab environments.}
    \label{tab:env_prompts}
    \begin{tabular}{l|c}
    \toprule
        Environment & Prompts \\
    \midrule
     \multirow{7}{*}{Avalon} & a screenshot of \\
     & a screenshot of a \\
     & a screenshot of many \\
      & a biome containing \\
     & a biome containing a \\
     & a biome containing many \\
     & a biome full of \\
     \midrule
     \multirow{6}{*}{Psychlab} & a render of \\
     & a render of a \\
     & a screenshot of \\
      & a screenshot of a \\
     & a screen showing \\
     & a screen showing a \\
    \bottomrule
    \end{tabular}
\end{table}

\section{Qualitative analyses}
\label{app:qualitiative}

\begin{figure}
    \centering
    \includegraphics[width=\textwidth]{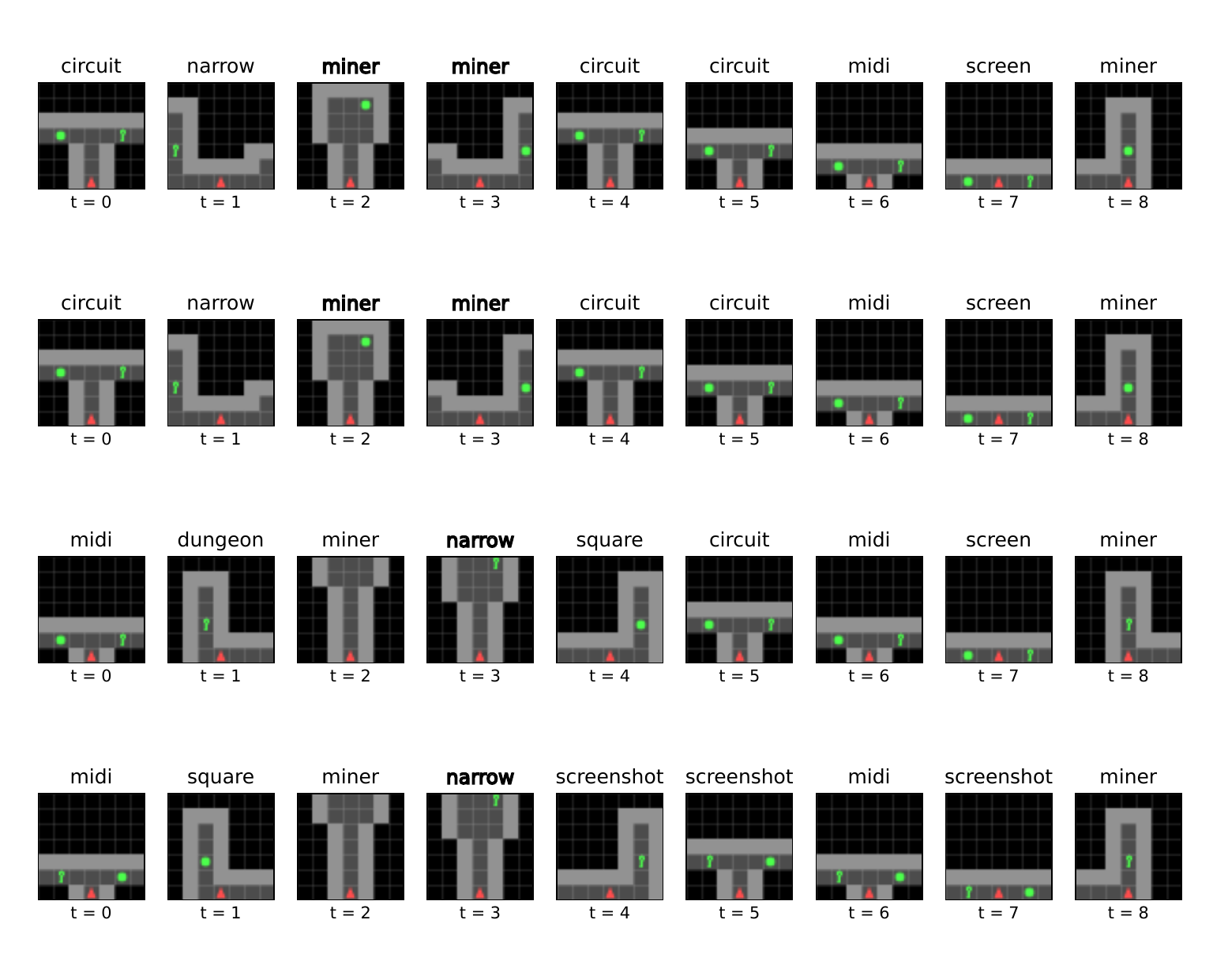}
    \caption{Episodes sampled for a trained SHELM policy on the MiniGrid-MemoryS11-v0 environment. The object \emph{ball} is consistently mapped to the token \emph{miner}, while the object \emph{key} maps to the token \emph{narrow}.}
    \label{fig:mem_episode}
\end{figure}

We show token mappings for the memory mechanism of SHELM to identify potential failures and successes.
\cref{fig:mem_episode} shows a few sample episodes from a trained policy on the Memory environment.
Clearly, CLIP is not capable of extracting semantics of the abstract 2D gridworld environments.
Thereby, it maps the ball to the token \emph{miner} and the key to the token \emph{narrow}.
For minimalistic 3D environments, CLIP is capable of extracting colors and shapes of objects as can be seen in \cref{fig:clip_backbone_miniworld}.
However, these results are still uninspiring since the token retrievals are very noisy.

\begin{figure}
    \centering
    \includegraphics[width=\textwidth]{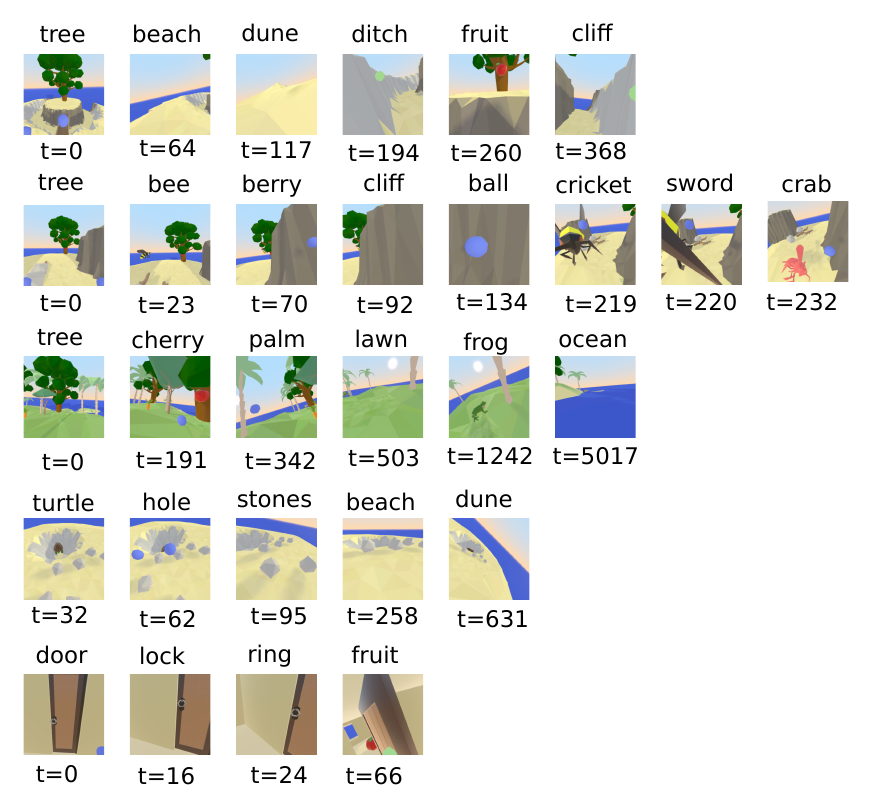}
    \caption{Episodes sampled from a random policy on various Avalon tasks.}
    \label{fig:avalon_episodes_appendix}
\end{figure}

We also visualize episodes collected by a random policy from the Avalon environment in \cref{fig:avalon_episodes_appendix}.
As opposed to the minimalistic MiniGrid and MiniWorld environments, CLIP successfully extracts semantically meaningful concepts which are subsequently passed to the memory.
There are still some failure cases though.
For example, in the second episode the agent is attacked by a bee.
The feelers of the bee are mistaken for a \emph{sword} as the bee moves into close proximity of the agent.
Further after the agent successfully defends itself against the bee, it mistakes the dead bee for a \emph{crab}.
Alleviating such failure cases would require grounding CLIP to the respective environment as was done in \citep{fan_minedojo_2022}.

\begin{figure}
    \centering
    \includegraphics[width=.9\textwidth]{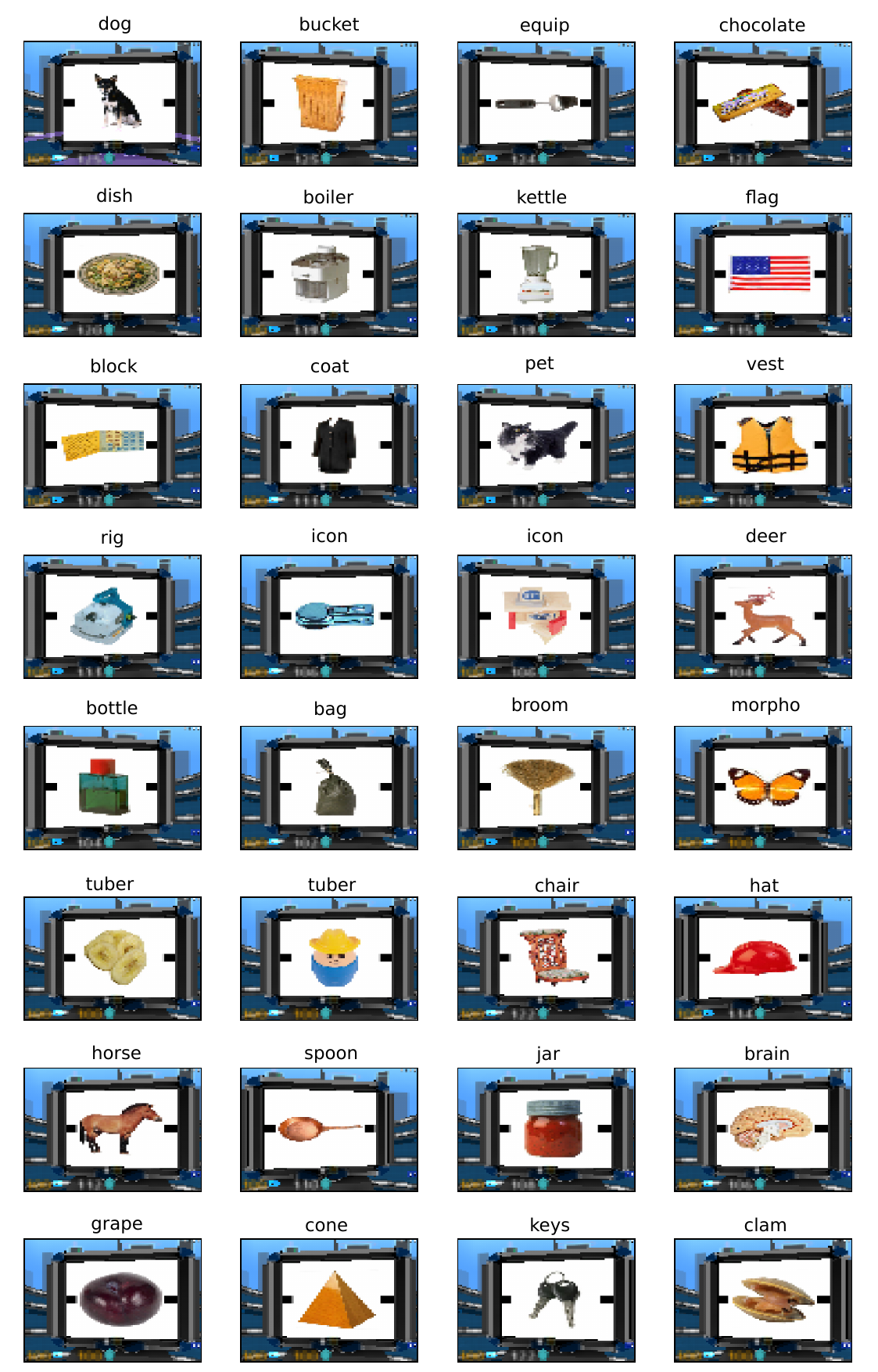}
    \caption{Sample observations for continuous recognition task of Psychlab. The agent must swipe in a certain direction depending on whether it has encountered an object before in the same episode.}
    \label{fig:dmlab_samples}
\end{figure}

\begin{figure}
    \centering
    \includegraphics[width=\textwidth]{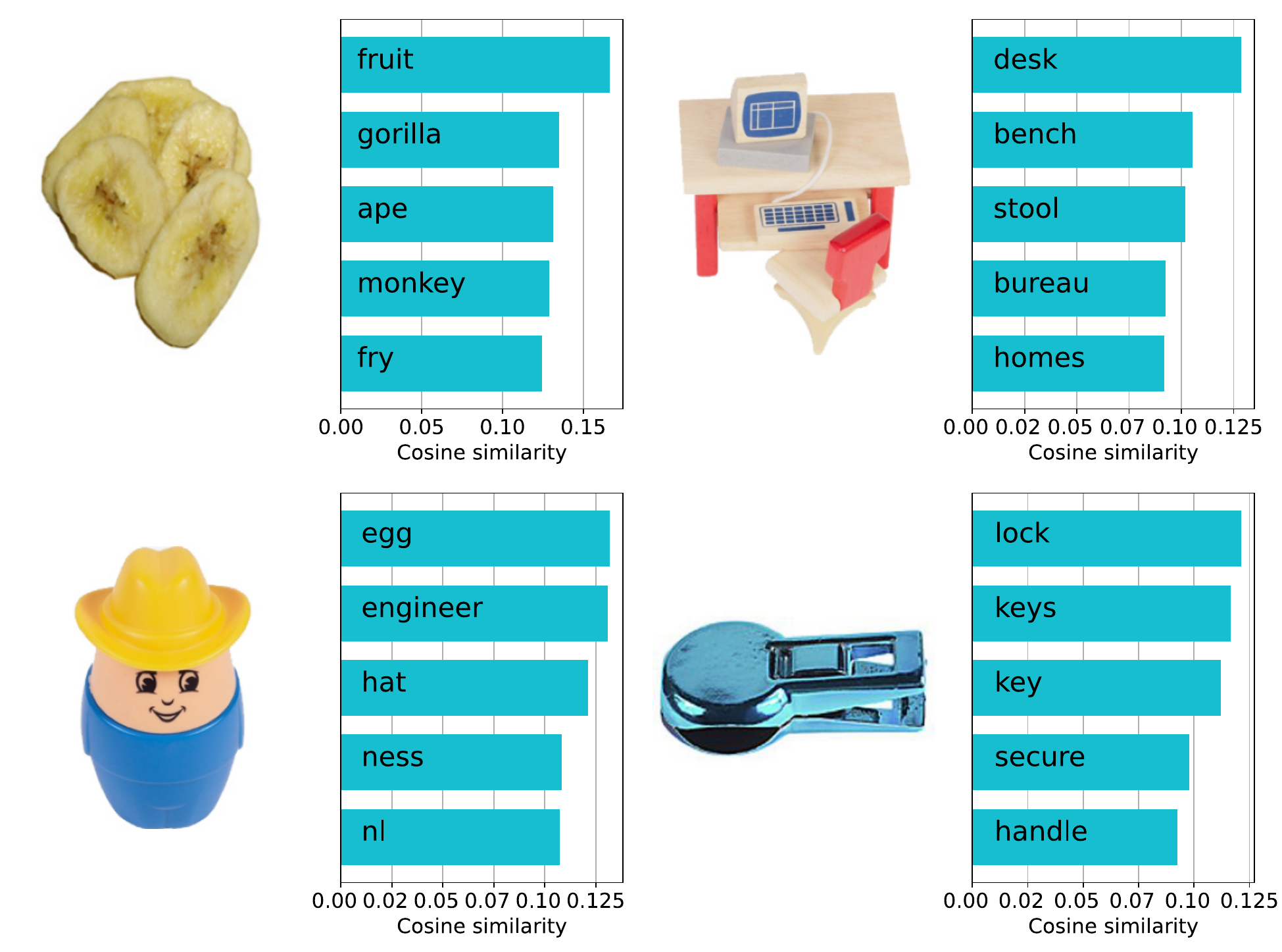}
    \caption{Top-5 token retrievals for objects that are conflated by SHELM at a lower resolution. For higher resolution CLIP successfully maps the objects to different tokens.}
    \label{fig:dmlab_corrects}
\end{figure}

Further, we visualize token mappings for objects the agent encounters in the continuous recognition task of Psychlab in \cref{fig:dmlab_samples}.
The majority of token retrievals are semantically correct. 
However, sometimes the agent confuses different objects or conflates to the same token for different objects.
An example for that are two middle objects in row 4 which are both mapped to the token \emph{icon}, or the first two objects in row 6 that are mapped to the token \emph{tuber}.
We suspect that this occurs due to downscaling to a lower resolution which is conducted within the DMLab engine.
Indeed, when taking a closer look at token retrievals at a higher resolution, they are mapped to different tokens (see \cref{fig:dmlab_corrects}).
Therefore, we consider two different aspects on how to alleviate this issue, (i) increasing the resolution of the observations and (ii) retrieving more than one token for an observation.
The former results in increased computational complexity, which we aim to avoid, since the task is computationally very expensive already.
The latter assumes that the retrieval differs at least in their top-$k$ most likely tokens and is a viable solution.

\section{Additional results}
\label{app:additional_results}

We show results for PPO, HELM, HELMv2, and SHELM on all Avalon tasks after 10\,M interaction steps in \cref{tab:avalon_hns_10m}.
Dreamerv2 attains the highest score on average after 10\,M timesteps, but performs on-par with SHELM. 
Further, there is no significant difference between the HELM variants though.
Since SHELM is demonstrably capable of extracting semantics from Avalon environments (see \cref{fig:avalon_episodes_appendix}), we believe that the main reason for these results is that the Avalon tasks do not require the agent to use its memory. 

\begin{table}[t]
\caption{Mean human normalized score for for all Avalon tasks for PPO, HELMv2, SHELM, and Dreamerv2 after 10M interaction steps. We show mean and standard deviations.}
\label{tab:avalon_hns_10m}
\begin{center}
\begin{tabular}{l c c c c c}
\toprule
Task & PPO & HELM & HELMv2 & SHELM & Dreamerv2 \\
\midrule
eat & 0.700 $\pm$ 0.070 & 0.671 $\pm$ 0.071 & 0.714 $\pm$ 0.063 & 0.693 $\pm$ 0.071 & 0.656 $\pm$ 0.077\\
move & 0.312 $\pm$ 0.066 & 0.277 $\pm$ 0.068 & 0.294 $\pm$ 0.061 & 0.291 $\pm$ 0.063 & 0.362 $\pm$ 0.073\\ 
jump & 0.232 $\pm$ 0.053 & 0.232 $\pm$ 0.058 & 0.219 $\pm$ 0.051 &  0.217 $\pm$ 0.050 & 0.233 $\pm$ 0.055\\
climb & 0.120 $\pm$ 0.042 & 0.125 $\pm$ 0.039 & 0.211 $\pm$ 0.052 &  0.118 $\pm$ 0.038 & 0.140 $\pm$ 0.048\\ 
descend & 0.203 $\pm$ 0.048 & 0.184 $\pm$ 0.044 & 0.108 $\pm$ 0.035 & 0.202 $\pm$ 0.048 & 0.283 $\pm$ 0.089 \\ 
scramble & 0.241 $\pm$ 0.056 & 0.213 $\pm$ 0.052 & 0.301 $\pm$ 0.056 &  0.271 $\pm$ 0.058 & 0.306 $\pm$ 0.089\\ 
stack & 0.094 $\pm$ 0.035 & 0.058 $\pm$ 0.029 & 0.075 $\pm$ 0.031 & 0.115 $\pm$ 0.041 & 0.106 $\pm$ 0.041 \\ 
bridge & 0.020 $\pm$ 0.017 & 0.046 $\pm$ 0.026 & 0.040 $\pm$ 0.024 & 0.028 $\pm$ 0.018 & 0.046 $\pm$ 0.031 \\ 
push & 0.109 $\pm$ 0.039 & 0.069 $\pm$ 0.030 & 0.069 $\pm$ 0.029 & 0.110 $\pm$ 0.044 & 0.102 $\pm$ 0.044\\ 
throw & 0.000 $\pm$ 0.000 & 0.017 $\pm$ 0.020 & 0.000 $\pm$ 0.000 & 0.000 $\pm$ 0.000 & 0.007 $\pm$ 0.011 \\ 
hunt & 0.059 $\pm$ 0.029 & 0.077 $\pm$ 0.033 & 0.091 $\pm$ 0.034 & 0.091 $\pm$ 0.039 & 0.067 $\pm$ 0.034 \\ 
fight & 0.179 $\pm$ 0.046 & 0.189 $\pm$ 0.048 & 0.207 $\pm$ 0.062 & 0.185 $\pm$ 0.051 & 0.299 $\pm$ 0.070 \\ 
avoid & 0.234 $\pm$ 0.048 & 0.192 $\pm$ 0.047 & 0.211 $\pm$ 0.045 & 0.227 $\pm$ 0.050 & 0.377 $\pm$ 0.062 \\ 
explore & 0.172 $\pm$ 0.045 & 0.194 $\pm$ 0.051 & 0.191 $\pm$ 0.050 & 0.204 $\pm$ 0.052 & 0.196 $\pm$ 0.051 \\ 
open & 0.086 $\pm$ 0.035 & 0.134 $\pm$ 0.042 & 0.040 $\pm$ 0.024 & 0.102 $\pm$ 0.038 & 0.055 $\pm$ 0.031 \\ 
carry & 0.080 $\pm$ 0.031 & 0.101 $\pm$ 0.041 & 0.073 $\pm$ 0.030 & 0.054 $\pm$ 0.027 & 0.087 $\pm$ 0.035 \\ 
\midrule
navigate & 0.006 $\pm$ 0.007 & 0.000 $\pm$ 0.000 & 0.000 $\pm$ 0.003 & 0.006 $\pm$ 0.008 & 0.006 $\pm$ 0.008\\ 
find & 0.005 $\pm$ 0.007 & 0.000 $\pm$ 0.000 & 0.000 $\pm$ 0.000 &0.000 $\pm$ 0.000 & 0.003 $\pm$ 0.004\\ 
survive & 0.049 $\pm$ 0.015 & 0.066 $\pm$ 0.005 & 0.041 $\pm$ 0.013 & 0.051 $\pm$ 0.016 & 0.033 $\pm$ 0.012 \\ 
gather & 0.007 $\pm$ 0.006 & 0.008 $\pm$ 0.019 & 0.009 $\pm$ 0.007 & 0.006 $\pm$ 0.003 & 0.006 $\pm$ 0.004\\ 
\midrule
overall & 0.146 $\pm$ 0.010 & 0.143 $\pm$ 0.010 & 0.145 $\pm$ 0.010 & 0.155 $\pm$ 0.011 & 0.168 $\pm$ 0.012 \\ 
\bottomrule
\end{tabular}
\end{center}
\end{table}

To further investigate this finding we run the memory-less PPO baseline for 50\,M interaction steps and compare it to results reported in \citet{albrecht_avalon_2022} in \cref{tab:avalon_50m}.
Surprisingly, our PPO baseline significantly outperforms the PPO baseline of \citet{albrecht_avalon_2022}.
Further, we find that our PPO baseline performs on-par with all other memory-based baselines, yielding further evidence that a memory mechanism is not imperative to solve the Avalon tasks.

\begin{table}
\caption{Mean human normalized score over all Avalon tasks after 50M interaction steps. We show mean and standard deviations. Results for PPO, Dreamerv2, and Impala were taken from \citet{albrecht_avalon_2022}. Asterisk indicates our own results.}
\label{tab:avalon_50m}
    \centering
    \begin{tabular}{cc}
    \toprule
    Method & Mean Human Normalized Score \\
    \midrule
    PPO & 0.165 $\pm$ 0.014 \\
    Dreamerv2 & 0.199 $\pm$ 0.012\\
    Impala & 0.203 $\pm$ 0.013 \\
    \midrule
    PPO* & 0.200 $\pm$ 0.012 \\
    \bottomrule
\end{tabular}
\end{table}

\section{Additional ablation studies}
\label{app:ablations}

\begin{figure}
    \centering
    \includegraphics[width=.5\textwidth]{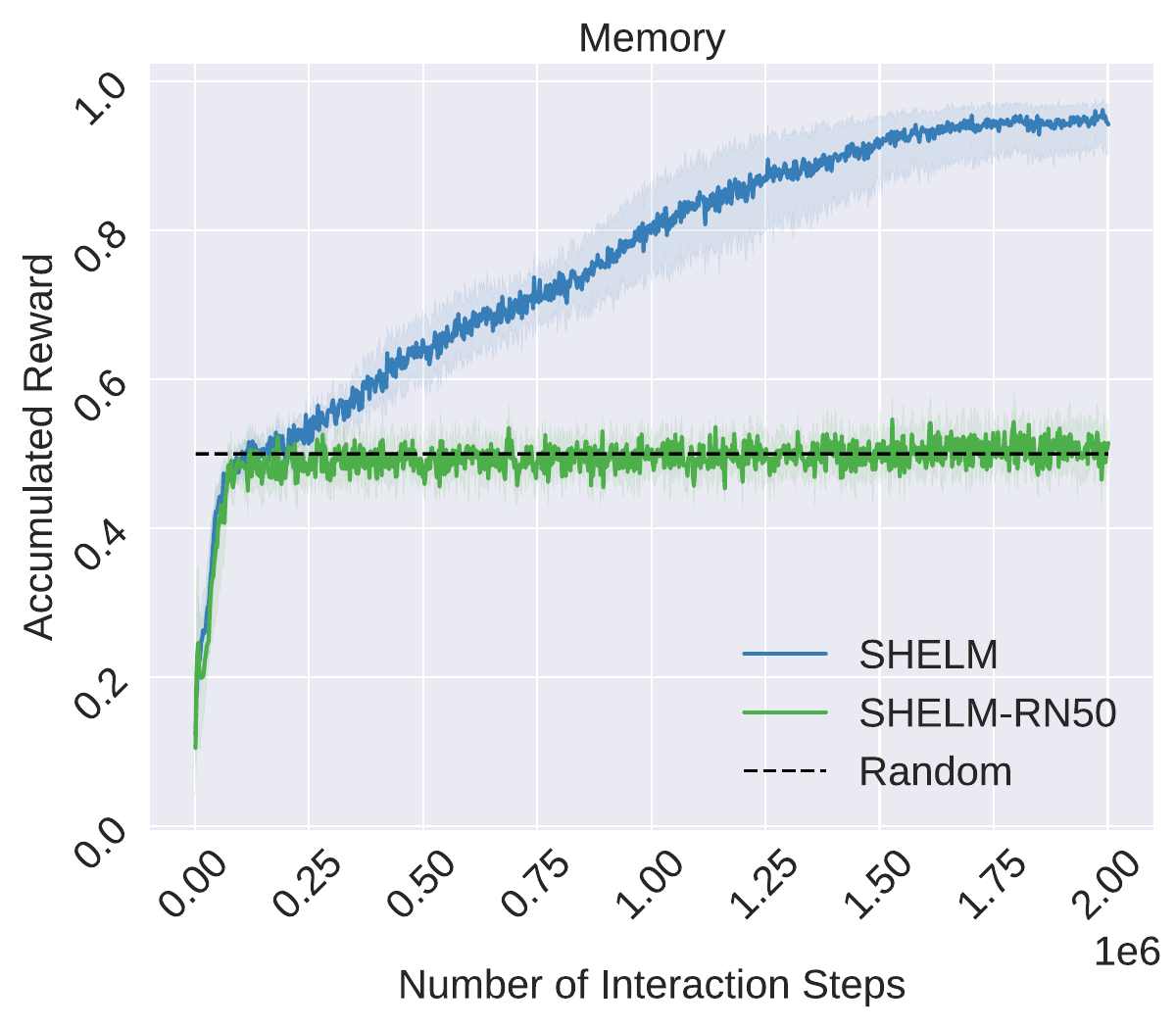}
    \caption{Ablation study on the choice of vision encoder used for the retrieval from our semantic database. We report IQM and 95\% CIs across 30 seeds on the MiniGrid-Memory task.}
    \label{fig:ablation_rn50}
\end{figure}

We provide an additional ablation study in the choice of vision encoder for SHELM.
Our qualitative results in \cref{subsec:extract_semantics} indicated that vision transformer based encoders are better in extracting semantics of synthetic environments.
To corroborate this finding, we run an experiment on the MiniGrid-Memory environment where we exchange the ViT-B/16 with a ResNet-50.
The results can be observed in \cref{fig:ablation_rn50}.
We observe that the ViT-B/16 encoder is much better in discriminating between the different objects which results in a substantial improvement over the RN50 encoder.

\section{Hyperparameters and training details}
\label{sec:hyperparams}

Since the memory component of SHELM does not need not be trained, our memory requirements are comparably low.
All our experiments were run on either a single GTX1080Ti or a single A100 GPU.
The time requirements of our experiments vary for each environment.
For MiniGrid and MiniWorld one run takes approximately two hours, while for Psychlab one run requires two days.
These experiments were run on a single GTX1080Ti.
For Avalon, we used a single A100 for training where one run to train for 10\,M interaction steps takes approximately 15 hours.

\paragraph{MiniGrid and MiniWorld}
We adapt the hyperparameter search from \citet{paischer_history_2022}.
Particularly, we search for learning rate in $\{5\text{e-}4, 3\text{e-}4, 1\text{e-}5, 5\text{e-}5\}$, entropy coefficient in $\{0.05, 0.01, 0.005, 0.001\}$, rollout length in $\{32, 64, 128\}$ for SHELM.
To decrease wall-clock time of HELM variants, we vary the size of the memory register of TrXL such that it can fit the maximum episode length.
We lower the number of interaction steps for the gridsearch if we observe convergence before the 500k interaction steps.   
If no convergence is observed within the 500K interaction steps, we tune for the entire duration.
We apply the same scheme for tuning the LSTM baseline and tune the same hyperparameters as in \citet{paischer_history_2022}.

\paragraph{Avalon}
After visual inspection of token retrievals for observations we found that there is no substantial difference in retrieved tokens for observations in close proximity to each other.
Therefore, we introduce an additional hyperparemeter, namely \textit{history-frameskip}.
Note that the history-frameskip is not functionally equivalent to the commonly used frameskip in, e.g., \citep{mnih_humanlevel_2015}. 
Rather, we actually discard frames within the skip.
For example, for a history-frameskip of 10 the agent only observes the first and the eleventh frame in the history branch.
The current branch observes every timestep as usual.
We search over history-frameskip in $\{ 3, 5, 10 \}$ and adapt the memory of the agent to $\{ 256, 128, 64 \}$ timesteps respectively.
Further we search over learning rate in $\{ 2.5\text{e-}4, 1\text{e-}4, 7\text{e-}5 \}$, and the number of retrieved tokens in $\{ 1, 2, 4 \}$.
If an observation is represented as more than one token, this effectively reduces the number of observations that fit into the memory register of TrXL, and thereby introduces a recency bias.
Hyperparameters for the Dreamerv2 baseline are taken from \citep{albrecht_avalon_2022}.
For the memory-less baseline we search over learning rates $\{ 5\text{e-}4, 2.5\text{e-}4, 1\text{e-}4, 5\text{e-}5 \}$.
We used their respective codebase to run our experiments.\footnote{https://github.com/Avalon-Benchmark/avalon}

\paragraph{PsychLab}
Due to the computational complexity of the Psychlab environments we only run a gridsearch over the learning rate in $\{ 5\text{e-}4, 3\text{e-}4, 1\text{e-}4, 5\text{e-}5 \}$. Further we use 64 actors and set the rollout size to 256. For SHELM on continuous-recognition we only retrieve the closest token for an observation. 

\section{Potential negative societal impact}
Our method relies on the use of FMs, which are trained on non-curated datasets which were crawled from the web. 
Therefore, these models readily reflect the biases and prejudices found on the web and, consequently, so might the resulting RL agent trained by our method. 
Generally, deployment of RL agents in the real world requires a carefully designed interface to guarantee safe  execution of selected actions.
That said, we do not expect any negative societal impact from our memory mechanism.
Our semantic memory enhances memory-based methods, thus, can yield more insights into why agents make certain decisions.

\section{Reproducibility Statement}
We make all our code and random seeds used in our experiments, as well as obtained results publicly available at \url{https://github.com/ml-jku/helm}.
The pretrained language encoder, as well as the pretrained CLIP encoder are publicly available on the huggingface hub \citep{wolf_transformers_2020}.

\end{document}